\documentclass[10pt, conference, compsocconf]{IEEEtran}

\newtheorem{thm}{Theorem}[section]

\newtheorem{problem}{Problem}
\newtheorem{exmp}{Example}

\newcounter{mycounter}

\makeatletter
\renewcommand*{\@opargbegintheorem}[3]{\trivlist
  \item[\hskip \labelsep{\it\quad  #1\ #2:}] {\it(#3)}\ }
\makeatother

\usepackage{stmaryrd}
\usepackage{graphicx} 
\usepackage{bbm}
\usepackage[final]{hyperref}
\usepackage{epsfig} 
\usepackage{amsmath}
\usepackage{amssymb}
\usepackage{epstopdf}
\usepackage{cite}
\usepackage[noend,ruled,linesnumbered,resetcount]{algorithm2e}
\SetKwComment{Comment}{$\triangleright$\ } {}
\usepackage{multirow}
\usepackage{rotating}
\usepackage{subfigure}
\usepackage{color}
\usepackage{mysymbol}
\usepackage{url}

\usepackage{tabularx,ragged2e,booktabs}
\usepackage[normalem]{ulem}
\usepackage[utf8x]{inputenc}
\usepackage{multirow}

\newcommand{\RNum}[1]{\uppercase\expandafter{\romannumeral #1\relax}}

\usepackage[hang,flushmargin]{footmisc}
\makeatletter
\newcommand{\algorithmfootnote}[2][\footnotesize]{%
  \let\old@algocf@finish\@algocf@finish
  \def\@algocf@finish{\old@algocf@finish
    \leavevmode\rlap{\begin{minipage}{\linewidth}
    #1#2
    \end{minipage}}%
  }%
}
\makeatother

  \makeatletter
  \setbox0\hbox{$\xdef\scriptratio{\strip@pt\dimexpr
    \numexpr(\sf@size*65536)/\f@size sp}$}
\newcommand{\scriptveryshortarrow}[1][3pt]{{%
    \hbox{\rule[\scriptratio\dimexpr\fontdimen22\textfont2-.2pt\relax]
               {\scriptratio\dimexpr#1\relax}{\scriptratio\dimexpr.4pt\relax}}%
   \mkern-4mu\hbox{\let\f@size\sf@size\usefont{U}{lasy}{m}{n}\symbol{41}}}}

\makeatother

\IEEEoverridecommandlockouts

\begin{document}
%
\title{Transfer Reinforcement Learning under Unobserved Contextual Information}


\author{\IEEEauthorblockN{Yan Zhang $\quad$ and $\quad$ Michael M. Zavlanos \thanks{This work is supported in part by AFOSR under award \#FA9550-19-1-0169 and by NSF under award CNS-1932011.}}
\IEEEauthorblockA{\textit{Department of Mechanical Engineering and Materials Science, Duke University, USA}  \\
		$\{$yan.zhang2, michael.zavlanos$\}$@duke.edu}
	
}


\maketitle

\begin{abstract}
In this paper, we study a transfer reinforcement learning problem where the state transitions and rewards are affected by the environmental context. Specifically, we consider a demonstrator agent that has access to a context-aware policy and can generate transition and reward data based on that policy. These data constitute the experience of the demonstrator. Then, the goal is to transfer this experience, excluding the underlying contextual information, to a learner agent that does not have access to the environmental context, so that they can learn a control policy using fewer samples. It is well known that, disregarding the causal effect of the contextual information, can introduce bias in the transition and reward models estimated by the learner, resulting in a learned suboptimal policy. To address this challenge, in this paper, we develop a method to obtain causal bounds on the transition and reward functions using the demonstrator's data, which we then use to obtain causal bounds on the value functions. Using these value function bounds, we propose new $Q$ learning and UCB-$Q$ learning algorithms that converge to the true value function without bias. We provide numerical experiments for robot motion planning problems that validate the proposed value function bounds and demonstrate that the proposed algorithms can effectively make use of the data from the demonstrator to accelerate the learning process of the learner.
\end{abstract}


\section{Introduction}

Reinforcement learning (RL) methods have been widely used to solve sequential decision making problems in unknown stochastic environments \cite{sutton2018reinforcement}. Often, these methods require many samples from the environment to find an optimal control policy, which becomes impractical when they are used to control physical systems, such as vehicles or robots, for which sampling can be expensive. One solution to this problem is to use Transfer Learning (TL) \cite{taylor2009transfer}. The goal in TL is to use the experience gained from solving prior {\em source tasks} to learn a policy for a new related {\em target task} using much fewer samples than if this new policy was learned from the beginning. 

TL methods can be classified depending on what specific knowledge is transferred from the source tasks to the target task. For example, the methods proposed in \cite{taylor2007transfer,barreto2017successor,finn2017model,konidaris2012transfer} focus on transferring value functions, transition functions, and policy functions between the source and target tasks. Specifically, the key idea in \cite{taylor2007transfer} is to transfer the value functions from prior tasks to a new task to provide a good initialization for the learning algorithm. On the other hand, in \cite{barreto2017successor} it is assumed that the source and target tasks share the same transition functions and only differ in the rewards, so that the transition features learnt from the source tasks can be transferred to the new task. In \cite{finn2017model}, a meta learning method is proposed to find a good initial policy for a new task, using experience from prior source tasks, which can be adapted to obtain the optimal policy using only a few samples. Along these lines, \cite{konidaris2012transfer} assumes that the source and target tasks share the same hiearchical structure so that the skills learned to solve a low level problem can be directly transferred to the target task.

In contrast to the methods proposed in \cite{taylor2007transfer,barreto2017successor,finn2017model,konidaris2012transfer}, the approaches discussed  in \cite{lazaric2008transfer,lazaric2011transfer,taylor2008transferringModel,tirinzoni2018importance,tirinzoni2019transfer} directly transfer the transition and reward samples from the source tasks to the target task. As discussed in \cite{taylor2008transferringModel}, transferring samples directly is more general compared to transferring value functions, transition functions, and policy functions, since it does not impose any requirements on the specific RL algorithm used to solve the source tasks. The reason is that every RL algorithm relies on transition and reward samples collected from the environment. 
Specifically, the methods proposed in \cite{lazaric2008transfer,lazaric2011transfer,taylor2008transferringModel} augment the target task dataset with the samples from the source tasks so that fewer new samples are needed from the target task. Since the source and target tasks usually have different reward and transition distributions, transferring the source task data to the target task introduces bias in the esitmated value functions or policies \cite{lazaric2011transfer,rosenstein2005transfer}.
The methods proposed in \cite{tirinzoni2018importance,tirinzoni2019transfer} rely on importance sampling to correct the bias introduced by using samples from source tasks. Since unbiased importance weights can result in large variance, biased weights are usually employed in practice to reduce the variance. Nevertheless, the computation of importance weights is expensive and requires fittting regression models for the rewards and transitions, as shown in  \cite{tirinzoni2018importance,tirinzoni2019transfer}. This introduces additional bias when the chosen regression model does not perfectly fit the true reward and transition functions.

In this paper, we propose a new TL framework that, as in \cite{lazaric2008transfer,lazaric2011transfer,taylor2008transferringModel,tirinzoni2018importance,tirinzoni2019transfer}, relies on sample transfer between source and target tasks but, unlike \cite{lazaric2008transfer,lazaric2011transfer,taylor2008transferringModel,tirinzoni2018importance,tirinzoni2019transfer}, it does not introduce bias in the learnt value functions and policies.
Specifically, we assume the presence of context in the environment, and consider a contextual Markov Decision Process (MDP) \cite{hallak2015contextual}, where the transition and reward functions are affected by this context that is subject to a fixed distribution. Moreover, we assume a  demonstrator agent, who can observe the environmental context and can collect transition and reward samples using a context-aware policy. These samples, excluding the contextual information, are provided to a learner agent as the source task dataset. The target task is to let the learner agent, who cannot observe the context,  find the optimal context-unaware policy in the same contextual MDP environment. In this problem, different contexts can be thought of as different source tasks. Compared to \cite{lazaric2008transfer,lazaric2011transfer,taylor2008transferringModel,tirinzoni2018importance,tirinzoni2019transfer}, since the context is hidden, the problem considered here is more challenging because the learner cannot identify which source task the transition and reward samples belong to. Therefore, the methods in \cite{lazaric2008transfer,lazaric2011transfer,tirinzoni2018importance,tirinzoni2019transfer} cannot be directly applied. Furthermore, since the source datasets are provided without contextual information, transferring these data using model-based RL methods like \cite{taylor2008transferringModel} can cause significant bias. This is because hidden contexts make the reward and transition models non-identifiable, 
as it is well known in the causal inference literature \cite{pearl2009causality}. To remove bias from the estimated transition and reward models and enable the learner agent to learn an optimal policy using fewer new samples, we extend the discretization method developed in Section 8.2 in \cite{pearl2009causality} so that we can derive causal bounds on the transition and reward functions using the source dataset. Then, given these causal bounds, we formulate two optimization problems that can be efficiently solved to obtain upper and lower bounds on the value functions, respectively. Using these value function bounds, we develop Causal Bound Constrained $Q$ learning \cite{watkins1992q} and UCB-$Q$ learning \cite{azar2017minimax,jin2018q,dong2019q} algorithms that converge to the true value function without bias, unlike the methods in \cite{lazaric2008transfer,lazaric2011transfer,taylor2008transferringModel,tirinzoni2018importance,tirinzoni2019transfer}. 
This is because, the true value functions are shown to lie within the proposed causal bounds and, therefore, projection of the value function iterates on the intervals defined by these bounds does not affect convergence of the $Q$ learning and UCB-$Q$ learning methods which are known to be unbiased.
The role of the causal bound constraints is to prevent these learning algorithms from returning poor value function estimates or exploring unnecessary states at the beginning of the learning process, reducing in this way the sampling complexity of our method.

To the best of our knowledge, the most relevant work to the method proposed here is \cite{zhang2017transfer}. Specifically, in \cite{zhang2017transfer} a multi-arm bandit problem is considered with binary action and reward spaces. Moreover, a similar method to the one proposed here is employed to compute causal bounds on the expected rewards of the bandit machines, which are used to improve on the performance of the UCB algorithm~\cite{auer2002finite}. Compared to \cite{zhang2017transfer}, here we consider a
sequential decision making problem on arbitrary discrete and finite action and reward spaces, which is more general. Moreover, we assume that the transition functions are also affected by the context and propose additional causal bounds for them. Note that the learning problem for the target task considered in this paper can be also thought of as a Partial-Observable Markov Decision Process (POMDP) problem. Therefore, the learner can apply any RL algorithm used for POMDPs, e.g., \cite{jaakkola1995reinforcement,hausknecht2015deep}, to solve the problem by itself. Nevertheless, these methods are sample inefficient. Instead, here we use the source dataset given by the demonstrator to reduce the number of samples needed by the learner to find the optimal policy.

The rest of this paper is organized as follows. In Section~\ref{sec:Prelim}, we define the proposed TL problem and discuss modeling bias due to the hidden context at the learner agent.
In Section~\ref{sec:CausalBound_ValueFunctions}, we develop our proposed method to compute the causal bounds on the reward, transition, and value functions. In Section~\ref{sec:Algorithm}, we propose a $Q$ learning and a UCB-$Q$learning algorithm based on the value function bounds. 
In Section~\ref{sec:Exp}, we numerically validate the proposed TL algorithms and causal bounds on robot motion planning problems. Finally, in Section~\ref{sec:Conclusion}, we conclude the paper.

\section{Preliminaries and Problem Definition}
\label{sec:Prelim}
In this section, we provide formal definitions of Markov Decision Processes (MDP) and contextual MDPs. We also formulate the problem of transferring experience from contextual MDPs to standard MDPs where the context is not observable, and discuss the challenges that arise in doing so.

\subsection{MDPs and Contextual MDPs}
Consider a MDP defined as a $4-$tuple $(s_t, a_t,$ $P(s_{t+1} | s_t, a_t), R(s_t, a_t, s_{t+1}))$, where  $s_t \in \mathcal{S}$ and $a_t \in \mathcal{A}$ denote the state and action at time $t$, $P(s_{t+1} | s_t, a_t)$ is the transition probability from state $s_t$ to $s_{t+1}$ when taking action $a_t$, and $R(s_t, a_t, s_{t+1})$ is the reward received when action $a_t$ is taken at state $s_t$ to transition state $s_{t+1}$. In what follows, we assume that the state and action spaces, $\mathcal{S}$ and $\mathcal{A}$, respectively, are finite and discrete. Moreover, we define a policy function $\pi(a_t | s_t) \rightarrow [0,1]$ as the probability of choosing action $a_t$ at state $s_t$.
A run of the MDP of length $T$ that is generated according to policy $\pi$ is called an episode.
The goal is to find the optimal policy $\pi^\ast$ so that the discounted accumulated reward

\begin{equation*}
	\label{eqn:DiscountedAccumulatedReward}
	\mathbb{E}_{\rho^0} \big[ \sum_{t = 0}^{T} \gamma^{t} R(s_t, a_t, s_{t+1}) \big]
\end{equation*}
is maximized, where $\gamma \in (0, 1)$ is the discount factor and $\rho^0$ is the distribution of the agent's initial state. Furthermore, we define the state-based value function at state $s$, $V(s) := \mathbb{E} \big[ \sum_{t = 0}^{T} \gamma^{t} R(s_t, a_t, s_{t+1}) | s_0 = s \big]$, and the action-based value function at state-action pair $(s,a)$, $Q(s, a) := \mathbb{E} \big[ \sum_{t = 0}^{T} \gamma^{t} R(s_t, a_t, s_{t+1}) | s_0 = s, a_0 = a \big]$. Finding the optimal policy $\pi^\ast$ is equivalent to computing the above two value functions under the optimal policy.

Given a MDP, a contextual MDP is defined as a $5-$tuple $(s_t, a_t, P^u(s_{t+1} | s_t, a_t), R^u(s_t, a_t, s_{t+1}),$ $\rho(u))$, with transition function $P^u(s_{t+1} | s_t, a_t)$ and reward function $R^u(s_t, a_t, s_{t+1})$ inherited from the original MDP but parameterized by a contextual variable $u \in \mathcal{U}$ \cite{hallak2015contextual}.
The context variable $u$ is sampled from a distribution $\rho(u)$ at the beginning of an episode and remains fixed for the duration of that episode.
The contextual parametrization of the transition probabilities and reward functions in a contextual MDP is motivated by many practical applications. For example, the motion of a robot under the same control action is affected by contextual information related to the environmental conditions, e.g., lighting and wind conditions. Similarly, a patient's health status under the same treatment is affected by contextual information related to the patient's profile, e.g., their age, gender, or weight.
Given a contextual MDP problem, the goal is to find a parameterized policy function $\pi^\ast(a_t | s_t, u)$ that maximizes the contextual accumulated reward
\begin{equation*}
\label{eqn:DiscountedAccumulatedReward_wtContext}
\mathbb{E}_{\rho^0} \big[ \sum_{t = 0}^{\infty} \gamma^{t} R^u(s_t, a_t, s_{t+1}) \big]
\end{equation*}
for every $u \in \mathcal{U}$.

\subsection{The Experience Transfer Problem}
\label{sec:ExperienceTransfer}
In what follows, we consider the problem of transferring the experience from a demonstrator agent who is aware of the contextual information included in the environment to another learner agent who does not have access to this information. By transferring experience we are able to use much fewer data to learn optimal control policies in RL for the agent with no contextual information.

Specifically, consider a contextual MDP $(s_t, a_t,$ $P^u(s_{t+1} | s_t, a_t), R^u(s_t, a_t, s_{t+1}),\rho(u))$ and assume that a demonstrator agent can observe the context $u$ and has knowledge of the optimal contextual policy $\pi^\ast(a_t | s_t, u)$. We model the experience of the demonstrator as state transition and reward pairs $(s_t, a_t, s_{t+1}, r_t)$ that can be obtained by executing the optimal policy $\pi^\ast(a_t | s_t, u)$ as follows. At the beginning of episode $k$, the environment randomly samples a contextual variable $u_k$ from the distribution $\rho(u)$ and a starting state $s_0$ from the distribution $\rho^0$. The context $u_k$ is revealed to the demonstrator. Then, the demonstrator executes the optimal policy $\pi^\ast(a_t | s_t, u_k)$ for $T$ time steps and records the state transitions and rewards $(s_t, a_t, s_{t+1}, r_t)$, excluding information about the context. Upon termination of this episode, a new episode is initialized and the process is repeated for $K$ episodes. In the end, the experience of the demonstrator is collected in the set $\{(s_t, a_t, s_{t+1}, r_t)\}_{t = 0 : KT}$, which we call the observational data. 
Note that we do not require the samples of the demonstrator to be provided in the form of trajectories. Moreover, samples from different contexts need not be differentiated from each other. Compared to \cite{lazaric2008transfer,lazaric2011transfer,tirinzoni2018importance,tirinzoni2019transfer} where the source samples should be arranged according to different tasks or trajectories, we only need to know the joint distribution $P(s_t, a_t, s_{t+1}, r_t)$ in the source samples, which imposes minimal restrictions on the source data. This is important in applications where the demonstrators are not allowed to directly operate in the environment but only respond to independent samples of state and context pairs.

Together with the demonstrator, we also consider a learner agent who interacts with the same contextual environment but cannot observe $u$. In this case, the goal of the learner is to find the optimal policy $\pi^\ast(a_t | s_t)$ that maximizes the accumulated reward averaged over the unobservable context $u$, that is, the learner agent solves the problem
 
\begin{equation}
	\label{eqn:DiscountedAccumulatedReward_wtAveragedContext}
	\max_{\pi(a_t | s_t)} \; \mathbb{E}_{\rho^0, \rho(u)} \big[ \sum_{t = 0}^{\infty} \gamma^{t} R^u(s_t, a_t, s_{t+1}) \big].
\end{equation}

Note that Problem~\eqref{eqn:DiscountedAccumulatedReward_wtAveragedContext} can be solved by the learner alone using any standard RL algorithm, e.g., Q learning. An optimal policy $\pi^\ast(a_t | s_t)$ that does not depend on the context $u$ exists because the context variable $u$ is subject to a stationary distribution $\rho(u)$ and its statistics can be marginalized to get new transition and reward functions that do not contain $u$. 
Nevertheless, this process typically requires a lot of data to learn the optimal policy. Our goal is to use experience from a context-aware demonstrator to help the learner find the optimal policy using much fewer data.
We formally state the problem of interest in this paper as follows.
\begin{problem}
	\vspace{2mm}
	\label{ProblemDefinition}
	(Experience Transfer with Hidden Context) Given the experience data set $\{(s_t, a_t, s_{t+1}, r_t)\}_{t = 0 : KT}$ of a context-aware demonstrator that excludes contextual information, design  learning algorithms for a context-unaware learner agent that use these data to find the optimal policy for problem~\eqref{eqn:DiscountedAccumulatedReward_wtAveragedContext} with only a small number of new data samples.
\end{problem}

\begin{figure}[t]
	\centering
	\subfigure[Demonstrator \label{fig:CausalGraphs_Demo}] 
	{\includegraphics[width = .4\columnwidth]{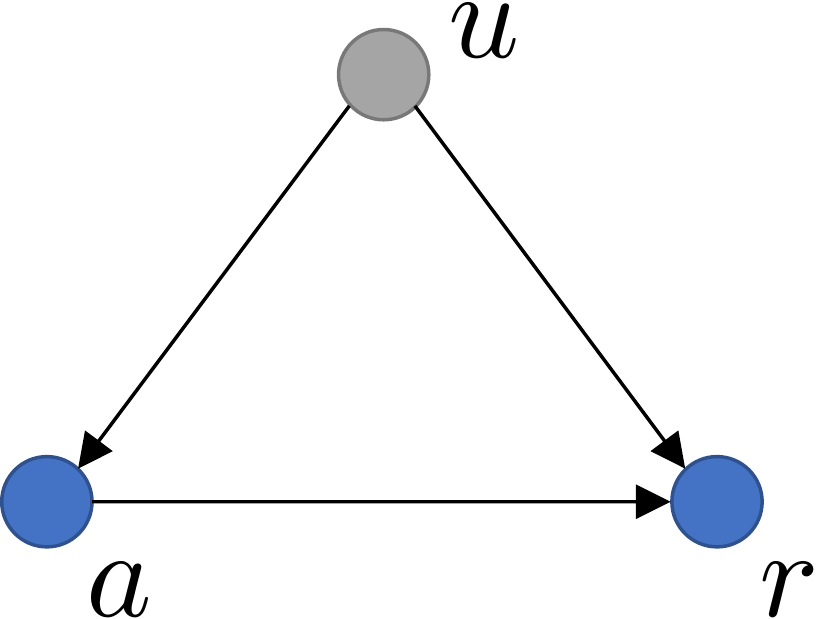}} 
	\subfigure[Learner 	\label{fig:CausalGraphs_Learner}]
	{\includegraphics[width = .4\columnwidth]{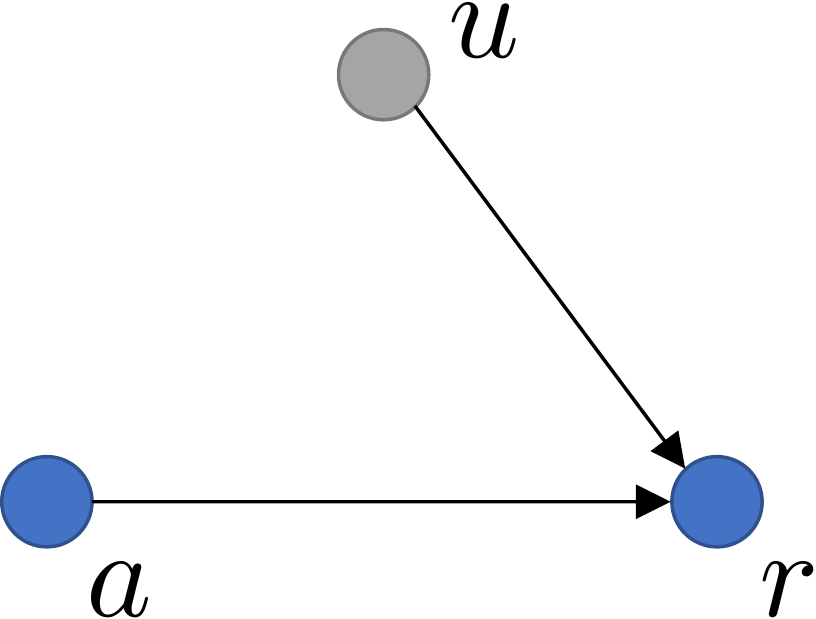}}
	\caption{Causal graphs to represent the behavior of the demonstrator and the learner. In both plots, the grey nodes represent the unobserved context $u$, and the blue nodes represent the action taken and the reward received at certain state. The arrow represents the direction of the causal effect. At each arrow, the end node's value is influenced by the source node's value. }
	\label{fig:CausalGraphs}
\end{figure}

The key challenge in solving Problem~\ref{ProblemDefinition} is that if the demonstrator's experience is used by the learner without considering the causal effect of the hidden context, it can result in biased estimation of the transition and reward models and, therefore, in a suboptimal learner policy. We discuss this issue in detail in Section~\ref{sec:ModelBias}. Note that there are many practical situations where the context that is known to the demonstrator is not revealed to the learner. An example is transferring motion planning experience to a robot that is not equipped with the necessary sensors to observe contextual information in its environment. Another example is transferring experience of medical treatment from one patient to another, where contextual information about the patients' profiles needs to remain confidential.

\subsection{Estimated Model Bias due to Hidden Context}
\label{sec:ModelBias}
A straightforward way to use the demonstrator's experience $\{(s_t, a_t, s_{t+1}, r_t)\}_{t = 0 : KT}$ is to directly estimate the model of the environment, that is, the transition and reward functions, from these samples. Then, any planning method for MDPs can be applied, e.g., Value Iteration \cite{sutton2018reinforcement}, to find the optimal policy. This approach is called model-based RL and has a lower sampling complexity compared to model-free RL approaches, see \cite{tu2018least,sun2019model}. However, when the data $\{(s_t, a_t, s_{t+1}, r_t)\}_{t = 0 : KT}$ is collected according to a hidden context, using this approach to estimate models of the transition and reward functions will result in bias and, therefore, a suboptimal policy for problem~\eqref{eqn:DiscountedAccumulatedReward_wtAveragedContext}.

To see this, consider the estimation of the expected reward $\mathbb{E}\big[r_t\big]$ by taking action $a_t$ at a certain state $s_t$. The causal relationship between the unobserved context $u$, the action $a$, and the reward $r$ for the demonstrator and the learner is shown in Figure~\ref{fig:CausalGraphs}. Specifically, in the case of the demonstrator, both the action and the reward are affected by the context $u$. On the other hand, in the case of the learner, the reward depends on the context $u$ but the action is independent of $u$ since context information is not observable at the learner. 
Assume now that the demonstrator collects the observational data as described in Section~\ref{sec:ExperienceTransfer}, which corresponds to the causal graph in Figure~\ref{fig:CausalGraphs_Demo}. Moreover, let $P(r, a) $ be the joint distribution of the action and  reward pair at state $s$, and  $P(s', a)$ be the joint distribution of the action and next state pair  at state $s$, which are computed using the observational data.
These distributions are called the observational distributions.
To correctly estimate the expected reward $\mathbb{E}\big[r_t\big]$ marginalizing out the unobserved context $u$, the learner needs to estimate the causal effect of action $a$,
\begin{equation}
\label{eqn:Estimation_DoX}
\mathbb{E} \big[ r | do(a) \big],
\end{equation}
where $do(a)$ assigns a specific value $a$ to the action regardless of what the context $u$ is, \cite{pearl2009causality}. Moreover, we have that $\mathbb{E} \big[ r | do(a) \big] = \sum_r r P(r | do(a))$, where

\begin{equation}
\label{eqn:Prob_doX}
P(r | do(a)) = \sum_u P(r | a, u) P(u).
\end{equation}
Since the context $u$ is missing from the observational distribution $P(r, a)$, we cannot compute \eqref{eqn:Prob_doX} directly. Instead, using $P(r, a)$, we can compute $\mathbb{E} \big[ r | a \big]$
as $\mathbb{E} \big[ r | a \big] = \sum_r r P(r | a) = \sum_r r \frac{P(r, a)}{P(a)} $. Furthermore,  we have that

\begin{equation}
\label{eqn:Prob_X}
\begin{split}
P(r | a) &= \frac{P(r, a)}{P(a)} = \frac{\sum_u P(r | a, u) P(a | u) P(u)}{\sum_u P(a | u) P(u)} \\
&= \sum_u \big[ P(r | a, u) (\frac{P(a | u) P(u)}{\sum_u P(a | u) P(u)}) \big].
\end{split}
\end{equation}
Comparing \eqref{eqn:Prob_X} and \eqref{eqn:Prob_doX}, we observe that $\mathbb{E} \big[ r | a \big] \neq \mathbb{E} \big[ r | do(a) \big]$. This is because the term $\frac{P(a | u) P(u)}{\sum_u P(a | u) P(u)}$ in \eqref{eqn:Prob_X} essentially reweights the distribution $P(u)$ in \eqref{eqn:Prob_doX} using $P(a|u)$, which is given by the demonstrator's policy function $\pi^\ast(a |s, u)$.
Therefore, the causal effect $\mathbb{E} \big[ r | do(a) \big]$ is non-identifiable when the context is missing from the observational distribution. And the same holds true for the transition probability $P(s' | s, do(a))$. Nevertheless, as discussed in \cite{pearl2009causality,zhang2017transfer}, it is possible to infer lower and upper bounds on both $\mathbb{E} \big[ r | do(a) \big]$ and $P(s' | s, do(a))$. In the next section, we show that such causal bounds can be computed given the observational data of the demonstrator and they can be used to help the learner learn faster.

\begin{figure}[t]
	\centering
	\includegraphics[width = .4\columnwidth]{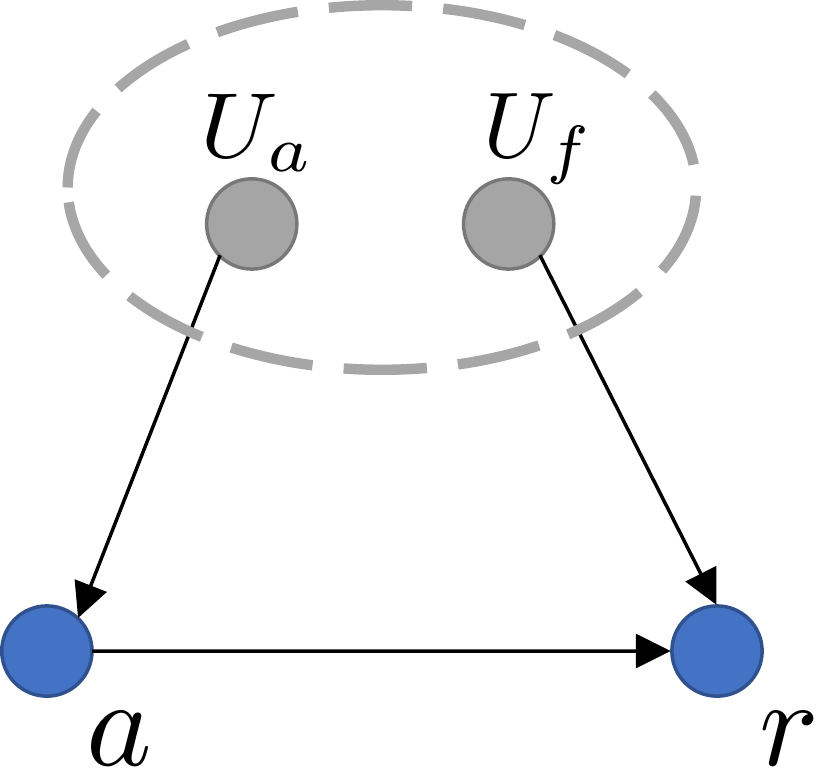}
	\caption{An auxillary causal graph with auxillary variables $U_a$ and $U_f$.}
	\label{fig:AuxillaryCausalGraph}
\end{figure}

\begin{table}[t]
	\centering
	\begin{tabular}{| c | c | c | c |} 
		\hline
		Index & a = 1 & a = 2 &  a = 3 \\
		\hline
		1 & 1 & 1 & 1 \\ 
		2 & 1 & 1 & 2 \\
		3 & 1 & 2 & 1 \\
		4 & 1 & 2 & 2 \\
		5 & 2 & 1 & 1 \\
		6 & 2 & 1 & 2 \\
		7 & 2 & 2 & 1 \\
		8 & 2 & 2 & 2 \\		
		\hline
	\end{tabular}
	\caption{\scriptsize All possible mappings from the action space $\mathcal{A} = \{1, 2, 3\}$ to the reward space $\mathcal{R} = \{1, 2\}$. }
	\label{table:Mapping}
	\vspace{-2mm}
\end{table}

\section{Causal Bounds on the Value Functions}
\label{sec:CausalBound_ValueFunctions}
In this section, we first show how to compute upper and lower bounds on the causal effect $\mathbb{E} \big[ r | do(a) \big]$ at state $s$ given the observational distribution $P(r, a)$. 
Our approach is a generalization of the method proposed in \cite{pearl2009causality,zhang2017transfer} to compute upper and lower bounds on the reward function using observational data for binary action and reward spaces to the case where these are any discrete and finite spaces. 
Then, we show how to extend this method to compute causal bounds on the transition probability $P(s' | s, a)$ for any transition $(s, a, s')$. 
Finally, we propose a way to obtain upper and lower bounds on the state-based optimal value function $V^\ast(s)$ and the action-based optimal value function $Q^\ast(s,a)$ for any state $s$ and action $a$, using the causal bounds on the rewards and transition probabilities obtained by the observational data.
Similar to \cite{zhang2017transfer}, throughout this section, we assume that the observational distributions $P(r, a)$ and $P(s', a)$ at state $s$ can be well estimated. This requires the demonstrator to take enough samples so that $P(r, a)$ and $P(s', a)$ are asymptotically correct according to the Law of large numbers.

\subsection{Causal Bounds on the Reward Function}
\label{sec:CausalBound_Reward}
Without loss of generality, let the action and reward spaces be indexed as $\mathcal{A} = \{1, 2, \dots, N_a\}$ and $\mathcal{R} =  \{1, 2, \dots, N_r\}$, respectively. For example, in a robot motion planning problem,  the actions indexed by $1$ and $2$ may correspond to 'go up' and 'go right'.
The causal effect of taking action $a$ is denoted as $\mathbb{E}\big[r | do(a) \big]$, where $\mathbb{E}\big[r | do(a) \big] = \sum_{r \in \mathcal{R}} r P \big( r | do(a) \big)$.
At state $s$, there exist $|\mathcal{R}|^{|\mathcal{A}|}$ mappings from the space $\mathcal{A}$ to $\mathcal{R}$. To see this, consider an example where $\mathcal{A} = \{1, 2, 3\}$ and $\mathcal{R} = \{1, 2\}$.  All the possible mappings are listed in Table~\ref{table:Mapping}. Mapping $3$ in Table~\ref{table:Mapping} means that actions $1$, $2$ and $3$ receive rewards $1$, $2$ and $1$, respectively. 

Next, we build an auxillary causal graph, similar to the one in Section 8.2 in \cite{pearl2009causality}, 
with two underlying random variables $U_a$ and $U_f$ as shown in Figure~\ref{fig:AuxillaryCausalGraph}. Let the supports of $U_a$ and $U_f$ be $\{1, 2, \dots, N_a\}$ and $\{1, 2, \dots, |\mathcal{R}|^{|\mathcal{A}|}\}$, respectively. Here, $U_a$ is the action choice, and $U_f$ is the action-to-reward mapping index. The causal relationship in Figure~\ref{fig:AuxillaryCausalGraph} is defined as
\begin{equation}
	\label{eqn:AuxillaryCausalRelaion}
	a = U_a \;\;\; \text{and } \;\;\;  r =f(a, U_f),
\end{equation}
where $f(a, U_f)$ returns the reward at entry $a$ in the mapping $U_f$. For instance, in Table~\ref{table:Mapping}, $f(3, 7) = 1$. Let also $q_{ij} = P(U_a = i, U_f = j)$.
As discussed in Section 8.2 in \cite{pearl2009causality}, we can construct the joint distribution $\{q_{ij}\}$ in the auxiliary graph in Figure~\ref{fig:AuxillaryCausalGraph} to reflect any possible causal relationship $P(r|do(a)))$ in the graph in Figure~\ref{fig:CausalGraphs_Demo}, regardless of the value of $U_a$. To do so, we need to construct $\{ q_{ij} \}$ as

\begin{equation}
	\label{eqn:CausalEffectProb_AuxillaryDistribution}
	P \big( r | do(a) \big) = \sum_{i \in  \{1, 2, \dots, N_a\}} \sum_{j \in \mathcal{S}_f(r, a)} q_{ij},
\end{equation}
where $\mathcal{S}_f(r, a)$ is an index set such that for every $j \in \mathcal{S}_f(r, a)$, $f(a, j) = r$. 
To see how the distribution $\{ q_{ij} \}$ that is constructed according to \eqref{eqn:CausalEffectProb_AuxillaryDistribution} generates the causal relationship $P \big( r | do(a) \big)$, consider the example in Table~\ref{table:Mapping}. Suppose the goal is to construct a distribution $\{q_{ij}\}$, with $i = {1, 2, 3}$ and $j = 1, 2, \dots, 8$, so that the auxillary graph can reflect the causal relationship $P(r = 1 | do(a = 2))$. Since the mappings $\{1, 2, 5, 6\}$ assign reward $r=1$ to action $a=2$, 
the probability $P(r = 1 | do(a = 2))$ consists of all events $U_f \in \{1, 2, 5, 6\}$. The value of $U_a$ is irrelevant since we fix the action to $do(a = 2)$. Therefore, we have that $P \big( r = 1 | do(a = 2) \big) = \sum_{i \in  \{1, 2, 3\}} \sum_{j \in \{1, 2, 5, 6\} } q_{ij}$.\\

Moreover, we can construct the joint distribution $\{q_{ij}\}$ in the auxillary graph in Figure~\ref{fig:AuxillaryCausalGraph} to reflect the observational distribution $P(r, a)$ in the graph in Figure~\ref{fig:CausalGraphs_Demo}. For this, we need to construct $\{q_{ij}\}$ so that

\begin{equation}
	\label{eqn:CausalEffectProb_ObservationalDistribution}
	P(r, a) = \sum_{j \in \mathcal{S}_f(r, a)} q_{aj}, \text{ for any } (r, a).
\end{equation}
The difference between \eqref{eqn:CausalEffectProb_ObservationalDistribution} and \eqref{eqn:CausalEffectProb_AuxillaryDistribution} is that \eqref{eqn:CausalEffectProb_ObservationalDistribution} requires that $U_a = a$. This is because the observational distribution $P(r, a)$ is generated using the auxillary graph in Figure~\ref{fig:AuxillaryCausalGraph}, where the action $a$ depends on $U_a$ according to the definition in \eqref{eqn:AuxillaryCausalRelaion}.
Using \eqref{eqn:CausalEffectProb_AuxillaryDistribution} and \eqref{eqn:CausalEffectProb_ObservationalDistribution}, and given the observation distribution $P(r, a)$ of the demonstrator, we can find an upper bound (or lower bound) on $\mathbb{E} \big[ r | do(a) \big]$ by solving the optimization problem
\begin{equation}
	\label{eqn:Optimization_CausalBound}
	\begin{split}
	& \max_{ \{q_{ij}\} } \; (\text{or} \min_{ \{q_{ij}\} } ) \; \sum_{r \in \mathcal{R}} r  \sum_{i \in  \{1, 2, \dots, N_a\}} \sum_{j \in \mathcal{S}_f(r, a)} q_{ij} \\
	& \text{s.t. } \sum_{j \in \mathcal{S}_f(r, a)} q_{aj} = P(r, a), \text{ for all } r \in \mathcal{R}, a \in  \mathcal{A}.
	\end{split}
\end{equation}
Note that problem~\eqref{eqn:Optimization_CausalBound} can be easily modified to incorporate  any prior knowledge on the causal relationship of the actions other than the action of interest, $P(r | do(m))$, $m \neq a$. For example, if $P(r | do(m))$ for action $m$ at state $s$ is known and the goal is to infer the causal effect of action $a \neq m$, we can add 
\begin{equation*}
	\label{eqn:AdditionalConstraint}
	\sum_{i \in  \{1, 2, \dots, N_a\}} \sum_{j \in \mathcal{S}_f(r, m)} q_{ij} = P(r | do(m))	
\end{equation*}
as additional constraints to problem~\eqref{eqn:Optimization_CausalBound} and obtain a tigher bound for action $a$.

{\begin{table}[t]
	\centering
	\begin{tabular}{| c | c | c | c |} 
		\hline
		Index & a = 1 & a = 2 \\
		\hline
		1 & 1 & 1 \\ 
		2 & 1 & 2 \\
		3 & 2 & 1 \\
		4 & 2 & 2 \\
		\hline
	\end{tabular}
	\caption{\scriptsize All possible mappings from the action space $\mathcal{A} = \{1, 2\}$ to the space of next possible state $\mathcal{S}' = \{1, 2\}$. }
	\label{table:Mapping_Transition}
	\vspace{-6mm}
\end{table}}

\subsection{Causal Bounds on the Transition Probabilities}
\label{sec:CausalBound_Transition}

We can directly extend the approach described in Section \ref{sec:CausalBound_Reward} to compute causal bounds on the state transition probabilities. At state $s$, let $P(a, s')$ be the joint distribution of taking action $a$ to reach state $s'$, which can be computed from the demonstrator's experience. Moreover, for any state $s$, let $\mathcal{S}' = \{1, 2, \dots, N_{s'}\}$ be the space of all possible next states. Then, similar to Section \ref{sec:CausalBound_Reward}, there exist $|\mathcal{S}'|^{|\mathcal{A}|}$ possible  mappings from the space $\mathcal{A}$ to $\mathcal{S}'$. For example, when the spaces of the actions and next states are $\mathcal{A} = \{1, 2\}$ and $\mathcal{S}' = \{1, 2\}$, the mappings are listed in Table~\ref{table:Mapping_Transition}. Let the random variable $U_{s'}$ select a mapping index from the set $\{1, 2, \dots, |\mathcal{S}'|^{|\mathcal{A}|} \}$.
We can construct the same auxillary causal graph as in Figure~\ref{fig:AuxillaryCausalGraph}, where the node $r$ is replaced by $s'$ and the causal relationship is defined as

\begin{equation*}
\label{eqn:AuxillaryCausalRelaion_Probability}
a = U_a \;\;\; \text{and } \;\;\;  s' =f_{s}(a, U_{s'}),
\end{equation*}
where $f_{s}(a, U_{s'})$ returns the state index at entry $a$ in the mapping $U_{s'}$. 
For example, in Table~\ref{table:Mapping_Transition}, we have that $f_s(1, 3) = 2$.
As before, let $q_{ij} = P(U_a = i, U_{s'} = j)$. Similar to \eqref{eqn:CausalEffectProb_AuxillaryDistribution}, we have that

\begin{equation}
\label{eqn:CausalEffectTransition_AuxillaryDistribution}
P \big( s' | do(a) \big) = \sum_{i \in  \{1, 2, \dots, N_a\}} \sum_{j \in \mathcal{S}_{f_s}(s', a)} q_{ij},
\end{equation}
where $\mathcal{S}_{f_s}(s', a)$ is an index set such that $f_s(a, j) = s'$ for every $j \in \mathcal{S}_{f_s}(s', a)$. 
Moreover, similar to \eqref{eqn:CausalEffectProb_ObservationalDistribution}, we have that

\begin{equation}
\label{eqn:CausalEffectTransition_ObservationalDistribution}
P(s', a) = \sum_{j \in \mathcal{S}_{f_s}(s', a)} q_{aj}.
\end{equation}
Then, given the observational distribution $P(a, s')$, we can find an upper bound (or lower bound) on $P(s' | do(a))$ by solving the optimization problem 

\begin{equation}
\label{eqn:Optimization_CausalBoundTransition}
\begin{split}
& \max_{ \{q_{ij}\} } \; (\text{or} \min_{ \{q_{ij}\} } ) \; \sum_{i \in  \{1, 2, \dots, N_a\}} \sum_{j \in \mathcal{S}_{f_s}(s', a)} q_{ij} \\
& \text{s.t. } \sum_{j \in \mathcal{S}_{f_s}(s', a)} q_{aj} = P(a, s'), \text{ for all } s' \in \mathcal{S}', a \in  \mathcal{A}.
\end{split}
\end{equation}
Same as in problem \eqref{eqn:Optimization_CausalBound}, here too any prior knowledge on $P(s' | do(m))$ corresponding to action $m$ at state $s$ can be modeled as additional constraints

\begin{equation*}
\label{eqn:AdditionalConstraint_Transition}
\sum_{i \in  \{1, 2, \dots, N_a\}} \sum_{j \in \mathcal{S}_{f_s}(s', m)} q_{ij} = P(s' | do(m))
\end{equation*}
that can be added to problem~\eqref{eqn:Optimization_CausalBoundTransition} so that a tighter bound can be achieved for action $a \neq m$.

\subsection{Causal Bounds on the Value Functions}
\label{sec:CausalBound_Value}
In this section, we discuss how to use the upper and lower bounds computed in Sections~\ref{sec:CausalBound_Reward} and \ref{sec:CausalBound_Transition} to find upper and lower bounds on the state-based and action-based functions $V(s)$ and $Q(s,a)$ for the learner agent.

First, we show how to compute such bounds for the function $V(s)$. For this, we rely on the linear programming formulation proposed in \cite{d1963probabilistic} to find the optimal value function $V(s)$ for a given discounted MDP problem

\begin{equation}
	\label{eqn:LP_ValueFunc}
	\begin{split}
		& \min_{ \{V(s)\} } \; \sum_{s \in \mathcal{S} } c(s) V(s) \\
		& \text{s.t. } V(s) \geq r(s, a) + \gamma \sum_{s'} P(s'|s, a) V(s'), \; \forall \;s, \;a.
	\end{split}
\end{equation}
where $c(s) > 0$ for all $s$ are called state-relevance weights. 
It is shown in \cite{d1963probabilistic} that the optimal value function $V(s)$ is the unique solution to problem~\eqref{eqn:LP_ValueFunc}. 

Solving problem \eqref{eqn:LP_ValueFunc} requires values for $r(s, a)$ and $P(s'|s, a)$, which can not be estimated without bias from the experience data of the demonstrator, as discussed in Section~\ref{sec:ModelBias}. For this reason, we instead utilize the bounds for $r(s,a)$ and $P(s'|s,a)$, denoted as $r(s, a) \in [\uline{r}(s, a), \overline{r}(s, a)]$ and $P(s' | s, a) \in [\uline{P}(s'|s, a), \overline{P}(s'|s, a) ]$, that are computed in Sections~\ref{sec:CausalBound_Reward} and \ref{sec:CausalBound_Transition}. 
Using these bounds, we can find an upper bound on the function $V(s)$ for every state $s$ by solving the optimization problem

\begin{equation}
	\label{eqn:LP_UpperBound}
		\begin{split}
		& \max_{ \substack{ r(s, a) \in \Gamma(s, a) \\ P(s'|s, a) \in \Xi(s, a, s') } } \; \min_{ \{V(s)\} } \; \sum_{s \in \mathcal{S} } c(s) V(s) \\
		& \text{s.t. } V(s) \geq r(s, a) + \gamma \sum_{s'} P(s'|s, a) V(s'), \; \forall \;s, \;a, 
	\end{split}
\end{equation}
where $\Gamma(s, a) := \{ r(s,a) \in [\uline{r}(s, a),\overline{r}(s, a)] \}$ and $\Xi(s, a, s'):=\{ P(s' | s, a) \in [\uline{P}(s'|s, a),\overline{P}(s'|s, a) ] \text{ and }$ $\sum_{s'} P(s'|s, a) = 1 \text{ for all } s \}$. Similarly, a lower bound on the function $V(s)$ for every $s$ can be found by the solution of the optimization problem

\begin{equation}
\label{eqn:LP_LowerBound}
\begin{split}
& \min_{ \substack{ r(s, a) \in \Gamma(s, a) \\ P(s'|s, a) \in \Xi(s, a, s') } } \; \min_{ \{V(s)\} } \; \sum_{s \in \mathcal{S} } c(s) V(s) \\
& \text{s.t. } V(s) \geq r(s, a) + \gamma \sum_{s'} P(s'|s, a) V(s'), \; \forall \;s, \;a.
\end{split}
\end{equation}

We can replace the inner minimization problem in \eqref{eqn:LP_UpperBound} with its dual form to obtain a maximization problem for \eqref{eqn:LP_UpperBound}.
\footnote{
The inner linear minimization problem in \eqref{eqn:LP_UpperBound} can be compactly written as $\min_{\vec{V} \in \mathbb{R}^{|\mathcal{S}|}} \;  \vec{c}^T \vec{V} \text{ s.t. }  A \vec{V} \geq \vec{b}$, 
where the vectors $\vec{c}, \vec{V}, \vec{b}$ and the matrix $A$ contains the variables $c(s)$, $V(s)$, $r(s, a)$ and $P(s'|s, a)$. Then, the corresponding dual problem takes the form 
$\max_{\vec{\Lambda} \in \mathbb{R}^{|\mathcal{S}| \times |\mathcal{A}|} } \; \vec{b}^T \vec{\Lambda} \text{ s.t. }  A^T \vec{\Lambda} = \vec{c}, \; \Lambda \geq 0$,
where the vector $\vec{\Lambda}$ contains the Lagrange multipliers for the inequality constraints in the primal problem~\eqref{eqn:LP_UpperBound}, \cite{bertsimas1997introduction}.
By \cite{d1963probabilistic}, the primal problem has an optimal solution and because of strong duality, the dual problem also has an optimal solution that has the same value with the primal problem, \cite{bertsimas1997introduction}.
}
Then, problems~\eqref{eqn:LP_UpperBound} and \eqref{eqn:LP_LowerBound} become quadratic constrained quadratic programs (QCQP) and can be solved using existing software packages, e.g., CPLEX.

Since problem~\eqref{eqn:LP_UpperBound} returns the reward and transition pair that maximizes the summation $\sum_{s \in \mathcal{S} } c(s) V(s)$ rather than the value function $V(s)$ at a specific state $s$, the solution of \eqref{eqn:LP_UpperBound} cannot be directly used as an upper bound on the value function $V(s)$ for any state $s$. The same holds true for the solution of problem~\eqref{eqn:LP_LowerBound} which can not be used as a lower bound on the function $V(s)$. Next, we discuss how to derive bounds on the value function $V(s)$ for a specific state $s$ given the solutions to problems~\eqref{eqn:LP_UpperBound} and \eqref{eqn:LP_LowerBound}.
\begin{thm}
\label{thm:Bd_ValueFunc}
	Assume the discrete and finite reward space is bounded, and let $\bar{R}$ and $\uline{R}$ be uniform upper and lower bounds on the reward functions $r(s, a)$. Moreover, let $\{ r^\ast, P^\ast, V^\ast(s) \}$ be the solution to problem~\eqref{eqn:LP_UpperBound}, and let $\bar{V}^\ast(s)$ be the true upper bound on the value at state $s$. Then, we have that
	
	\begin{equation}
	\label{eqn:UpBd_Value}
		\bar{V}^\ast(s) \leq V^\ast(s) + \frac{1}{c(s)} \big(\sum_{ \tilde{s} \neq s } c(\tilde{s}) V^\ast(\tilde{s})  - \frac{\uline{R}}{1 - \gamma} \sum_{ \tilde{s} \neq s } c(\tilde{s}) \big).
	\end{equation}
	Similarly, let $\{ r^\ast, P^\ast, V^\ast(s) \}$ be the solution to problem~\eqref{eqn:LP_LowerBound}, and let $\uline{V}^\ast(s)$ be the true lower bound on the value at state $s$. Then, we have that

	\begin{equation}
	\label{eqn:LwrBd_Value}
		\uline{V}^\ast(s) \geq V^\ast(s) - \frac{1}{c(s)} \big( \frac{\bar{R}}{1 - \gamma} \sum_{ \tilde{s} \neq s } c(\tilde{s}) - \sum_{ \tilde{s} \neq s } c(\tilde{s}) V^\ast(\tilde{s}) \big).
	\end{equation}
\end{thm}

The proof can be found in Appendix A. Theorem~\ref{thm:Bd_ValueFunc} implies that tight bounds on $V(s)$ can be obtained for any state $s$ by using \eqref{eqn:UpBd_Value} and \eqref{eqn:LwrBd_Value} and by selecting large enough $c(s)$ while fixing the $c(\tilde{s})$ for all $\tilde{s} \neq s$. 

Finally, denote the bounds on the value function $V(s)$ on the right hand side of \eqref{eqn:UpBd_Value} and \eqref{eqn:LwrBd_Value} as $\bar{V}(s)$ and $\uline{V}(s)$. Since the optimal action-based value function satisfies $Q(s, a) = r(s, a) + \gamma \sum_{s'} P(s'|s, a) V(s')$,  we can use the bounds $[\uline{V}(s), \bar{V}(s)]$ together with the bounds $[\uline{r}(s, a), \overline{r}(s, a)]$ and $[\uline{P}(s'|s, a), \overline{P}(s'|s, a) ]$ derived in Sections~\ref{sec:CausalBound_Reward} and \ref{sec:CausalBound_Transition} to obtain bounds
$[\uline{Q}(s, a), \overline{Q}(s, a) ]$ on the action-value function $Q(s,a)$, for all $s \in \mathcal{S}$ and $a \in \mathcal{A}$. This process is straighforward and is directly presented in Algorithm~\ref{alg:Q_bound} in Appendix B.

\section{Algorithm design}
\label{sec:Algorithm}
In this section, we present two approaches to combine the causal bounds $ [\uline{Q}(s, a), \overline{Q}(s, a) ]$ derived in Algorithm~\ref{alg:Q_bound} in Appendix~B with existing RL algorithms to accelerate the learning process of the learner. The first approach uses the causal bounds as the constraints of the value function in the $Q$ learning method \cite{watkins1992q}. The second approach uses the causal bounds to provide better exploration performance based on existing UCB-$Q$ exploration strategies in \cite{azar2017minimax,jin2018q,dong2019q}.

\subsection{Causal Bound Constrained $Q$ Learning}

Let the learner agent interact with the environment at state $s_t$ by taking action $a_t$, receiving reward $r(s_t, a_t)$, and observing the next state $s_{t+1}$. In standard $Q$ learning the agent's current action-value function is updated as
\begin{algorithm}[t]
	\small
	\caption{Causal Bound Constrained $Q$ learning}\label{alg:CBC-Q}
	\KwIn{Initial value function estimate $Q(s, a)$ for all $(s, a)$. Learning rate $\alpha_t$. Discount factor $\gamma$. Number of episodes $K$. Number of time steps in each episode $T$. Initial state distribution $\rho^0$. Context distribution $\rho(u)$. Causal bounds $[\uline{Q}(s_t, a_t), \overline{Q}(s_t, a_t) ]$ for all $(s, a)$. Parameter $\epsilon$ for the $\epsilon-$greedy policy \cite{sutton2018reinforcement}.}{
	\For{episode $1 \leq k \leq K$}{
		The learner samples the intial state $s_0$ and context $u_k$ from distribution $\rho^0$ and $\rho(u)$, respectively \;
		\For{time step $0 \leq t \leq T$}{
			The learner observes its state $s_t$ \;
			The learner takes action $a_t$ according to the $\epsilon$-greedy policy\;
			The learner receives reward $r(s_t, a_t)$ and observes its next state $s_{t+1}$ \;
			The learner updates its value function at $(s_t, a_t)$ according to \eqref{eqn:Qlearning_Constrained}.
		}
	}
	}
\end{algorithm}

\begin{equation}
	\label{eqn:Qlearning}
	\begin{split}
	Q(s_t, a_t) \leftarrow & (1 - \alpha_t)  Q(s_t, a_t)  \\
	& + \alpha_t \big( r(s_t, a_t) + \gamma  \max_a Q(s_{t+1}, a) \big),
	\end{split}
\end{equation}
where $\alpha_t$ is the learning rate. It is shown in \cite{watkins1992q} that $Q$ learning almost surely converges to the optimal value with sufficient exploration and an appropriate choice of the learning rate. Using the bounds $ [\uline{Q}(s, a), \overline{Q}(s, a) ]$ on the action-value function derived in Section~\ref{sec:CausalBound_Value} for the case that the context observed by the demonstrator is hidden to the learner, we can modify the above $Q$ learning update by introducing a projection operation as

\begin{equation}
\label{eqn:Qlearning_Constrained}
\begin{split}
Q(s_t, a_t) \leftarrow & \Pi_{[\uline{Q}(s_t, a_t), \overline{Q}(s_t, a_t) ]} \bigg( (1 - \alpha_t)  Q(s_t, a_t)  \\
& + \alpha_t \big( r(s_t, a_t) + \gamma  \max_a Q(s_{t+1}, a) \big) \bigg).
\end{split}
\end{equation}
We formally present the resulting Causal Bound Constrained $Q$ learning (CBC-$Q$ learning) method in Algorithm~\ref{alg:CBC-Q}. In the early learning phase of standard $Q$ learning in \eqref{eqn:Qlearning}, the value $Q(s,a)$ is incorrect for most state-action pairs.
By projecting onto the causal bounds in \eqref{eqn:Qlearning_Constrained}, the values of all state-action pairs are confined to a provably correct region that contains the optimal value, therefore accelerating the learning process. 
During the later stages of learning, the value iterates $Q(s,a)$ approach the true values, so that the causal bound constraints do not take effect anymore. Therefore, the update~\eqref{eqn:Qlearning_Constrained} is eventually reduced to the standard $Q$ learning update~\eqref{eqn:Qlearning}, which is unbiased \cite{watkins1992q}.

\subsection{Causal Bound Aided UCB-$Q$ Learning}

\begin{algorithm}[t]
	\small
	\caption{Causal Bound Aided UCB-$Q$ learning}\label{alg:CB-UCBQ}
	\KwIn{Initial upper confidential bound estimate $Q_U(s, a)$ for all $(s, a)$. Learning rate $\alpha_t$. Discount factor $\gamma$. Number of episodes $K$. Number of time steps in each episode $T$. Initial state distribution $\rho^0$. Context distribution $\rho(u)$. Causal bounds $[\uline{Q}(s_t, a_t), \overline{Q}(s_t, a_t) ]$ for all $(s, a)$.}{
		\For{episode $1 \leq k \leq K$}{
			Sample the intial state $s_0$ and context $u_k$ from distribution $\rho^0$ and $\rho(u)$, respectively \;
			\For{time step $0 \leq t \leq T$}{
				Let the agent observe its state $s_t$ \;
				Let the agent take action $a_t$ so that
				
				\begin{equation*}
				a_t = \arg \max_a \; Q_U(s_t, a) ;
				\end{equation*}
				The agent receives reward $r(s_t, a_t)$ and observes its next state $s_{t+1}$ \;
				The agent updates its upper confidential bound at $(s_t, a_t)$ according to \eqref{eqn:UCBQ} \;
				$Q_U(s_t, a_t) \leftarrow \min\big(Q_U(s_t, a_t), \overline{Q}(s_t, a_t)\big)$.
			}
		}
	}
\end{algorithm}

The convergence of $Q$ learning to the optimal value assumes that all states and actions are sufficiently sampled \cite{watkins1992q}. This requires the learner to follow a randomized policy function to explore the environment. In \cite{sutton2018reinforcement}, an $\epsilon-$greedy policy is discussed, which lets the agent choose the best action according to its current value function estimate with probability $1-\epsilon$ and choose other actions with probability $\epsilon$. However, this policy can be sample inefficient \cite{kearns2002near}. The UCB exploration strategy has been shown to improve on the sampling complexity of the $\epsilon-$greedy policy, first for Multi-Arm Bandit problems \cite{auer2002finite} and subsequently for general reinforcement learning problems \cite{jin2018q,azar2017minimax,dong2019q}. The key idea is that instead of estimating the value function, UCB methods keep an estimate of the upper confidential bound of the value function and choose the best action according to this upper confidential bound. Specifically, in \cite{jin2018q,dong2019q}, the upper confidential bound on the value function $Q_{U}(s, a)$ is updated as

\begin{equation}
	\label{eqn:UCBQ}
	\begin{split}
	Q_{U}(s_t, a_t) \leftarrow  (1 - \alpha_t) & Q_U(s_t, a_t) + \alpha_t \big( r(s_t, a_t)  \\
	& + \gamma  \max_a Q_U(s_{t+1}, a) + b_t\big),
	\end{split}
\end{equation}
where $b_t$ is an estimate of the confidential interval on the value function at $(s_t, a_t)$, which is usually of order $O(\frac{1}{\sqrt{k}})$ \cite{jin2018q,dong2019q} and $k$ is the number of times $(s_t, a_t)$ is visited at time $t$. Essentially, the more often every state-action pair $(s_t, a_t)$ is visited, the tighter the corresponding interval $b_t$ becomes \cite{jin2018q,dong2019q}. 
Given the bounds $[\uline{Q}(s, a), \overline{Q}(s, a) ]$ on the action-value function computed using Algorithm~\ref{alg:Q_bound} in Appendix~B, we add an additional step after update~\eqref{eqn:UCBQ}, i.e.,
\begin{equation}
	\label{eqn:UCBQ_constraint}
	Q_U(s_t, a_t) \leftarrow \min\big(Q_U(s_t, a_t), \overline{Q}(s_t, a_t)\big).
\end{equation}
We formally present the Causal Bound aided UCB-$Q$ learning method in Algorithm~\ref{alg:CB-UCBQ}. The key idea is that when the estimated upper confidential bound $Q_U(s, a)$ is higher than $\overline{Q}(s, a)$, Algorithm~\ref{alg:CB-UCBQ} uses the causal bound as the upper confidential bound. This prevents the upper confidential bound at some state-action pairs from being too optimistic, therefore, avoids unnecessary exploration and accelerates the learning process.

\section{Numerical experiments}
\label{sec:Exp}
In this section, we provide numerical results for TL in motion planning problems to validate the causal bounds presented in \eqref{eqn:UpBd_Value} and \eqref{eqn:LwrBd_Value} and the efficacy of Algorithms~\ref{alg:CBC-Q} and \ref{alg:CB-UCBQ} proposed in Section~\ref{sec:Algorithm}. 

\subsection{Unobserved Contextual Reward Function}
\label{sec:Sim_ContextualReward}
\begin{figure}[t]
	\centering
	\includegraphics[width = .45\columnwidth]{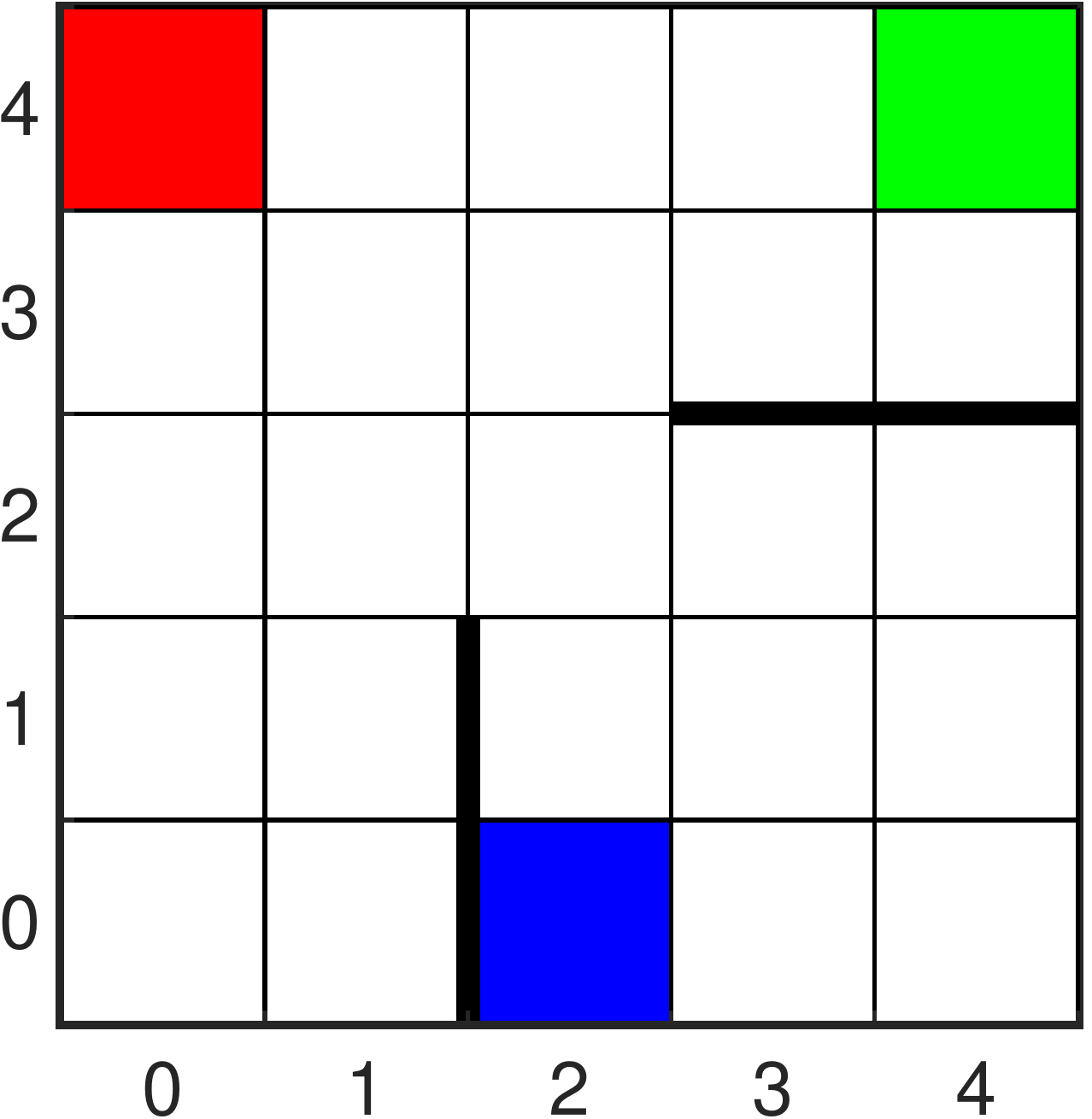}
	\caption{A motion planning problem where the agent starts from the blue cell and moves to the red or green cells. The rewards received when reaching the goals are affected by the context variable $u$.}
	\label{fig:field_Rwd}
	\vspace{-2mm}
\end{figure}

Consider the motion planning problem in Figure~\ref{fig:field_Rwd}. Let $s \in \{0, 1, \dots, 4\}^2$ denote the state (position) of the robot and $a \in \{1, 2, 3, 4\}$ denote its actions $\{\text{go up}, \text{go right},$ $\text{go down}, \text{go left}\}$. Let the transition be deterministic, that is, when we choose action $a = 1$ at $s = [2, 1]$, the robot transitions to $[2, 2]$, unless it collides with the walls in Figure~\ref{fig:field_Rwd}, in which case it remains at its current position. The red and green cells in Figure~\ref{fig:field_Rwd} are goal positions. When the robot takes action $a$ at state $s$ and the next position is not one of the two goal positions, it receives a reward $-1$. On the other hand, if the robot is already at the goal position, it remains there and receives a reward $0$ regardless of what action it takes. When the robot reaches a goal, it receives a reward according to the random context variable $u = \{0, 1\}$ in the environment. Specifically, when the robot reaches the red cell, if $u = 0$ (or $u = 1$), it receives a reward $10$ with probability $0.6$ (or $0.1$). Similarly, when the robot reaches the green cell, if $u=0$ (or $u = 1$), it receives a reward $5$ with probability $0.3$ (or $0.8$). The discount factor is $\gamma = 0.9$. The context variable is sampled from a Bernoulli distribution with parameter $0.8$.



In what follows, we focus on computing the causal bounds on the rewards at critical state-action pairs. Specifically, in this example we are interested in $([0, 3], 1)$, $([1, 4], 4)$, $([3, 4], 2)$ and $([4, 3], 1)$. 
In practice, given the data from the demonstrator, we can identify these critical state-action pairs either by expert knowledge on which state-action pairs are easily affected by potential hidden context, or by using a heuristic, that is, selecting those pairs that receive different rewards. In this example, only the state-action pairs above receive different rewards from the demonstrator's experience.  
Note that our approaches to compute the bounds on the value functions in \eqref{eqn:LP_UpperBound} and \eqref{eqn:LP_LowerBound} do not rely on the knowledge of these critical state-action pairs. 
\begin{table}[t]
	\centering
	\begin{tabular}{| c | c | c | c |} 
		\hline
		$(s, a)$ & $\mathbb{E}\big[r | s, do(a)\big]$ & $\mathbb{E}\big[r |s, a \big]$ & Causal bounds \\
		\hline
		$([0, 3], 1)$ & $1.2$ & $1.2$ & $\big[0.54, 3.84\big]$ \\
		$([1, 4], 4)$ & $1.2$ & $3.6$ & $\big[0.012, 8.5920\big]$ \\
		$([3, 4], 2)$ & $3.2$ & $1.8909$ & $\big[-0.3640, 4.3160\big]$ \\
		$([4, 3], 1)$ & $3.2$ & $3.2$ & $\big[1.94, 3.74\big]$ \\
		\hline
	\end{tabular}
	\caption{\scriptsize The true causal effects $\mathbb{E}\big[r | s, do(a)\big]$, direct estimate of the expected rewards from the observational distribution $P(r, a|s)$ without considering the hidden contexts and the causal bounds obtained by solving problem~\eqref{eqn:Optimization_CausalBound}. }
	\label{table:CausalBounds_Reward}
	\vspace{-6mm}
\end{table}

\begin{figure}[t]
	\centering
	\includegraphics[width = .65\columnwidth]{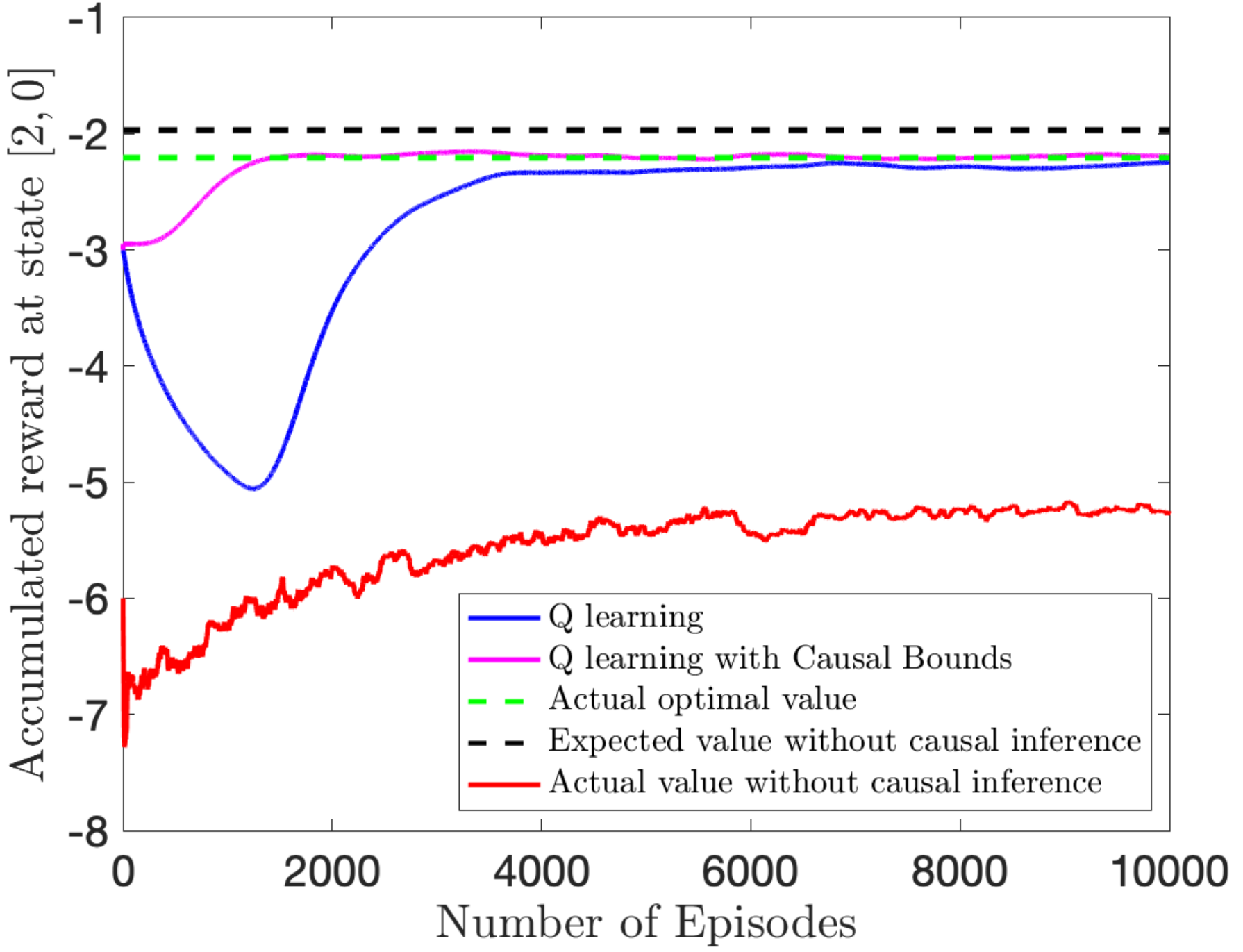}
	\caption{Comparison between the proposed Causal Bound constrained $Q$ learning and the standard $Q$ learning. The blue curve shows the learning progress of the $Q$ learning algorithm \cite{watkins1992q} without using data from the demonstrator. The magenta curve shows the performance of Algorithm~\ref{alg:CBC-Q}. Both curves are obtained by running these algorithms for $10$ trials and take the average. Meanwhile, the black (or green) dashed line is the optimal value at state $[2, 0]$ computed using $\mathbb{E}\big[r |s, a \big]$ (or $\mathbb{E}\big[r | s, do(a)\big]$ ). The red curve shows the evaluation of the sub-optimal policy computed using $\mathbb{E}\big[r |s, a \big]$.}
	\label{fig:Qlearning_wo_causal}	
	\vspace{-4mm}
\end{figure}

Consider the demonstrator agent who can observe the context variable $u$ and also knows the contextual optimal policy
\begin{equation}
	\label{eqn:OptimalPolicy_demo}
	\pi^\ast(s, u) =
	\begin{cases}
	& 4, \text{ when } s = [1, 4] \text{ and } u = 0,\\
	& 1, \text{ when } s = [1, 4] \text{ and } u = 1,\\
	& 4, \text{ when } s = [3, 4] \text{ and } u = 0,\\
	& 2, \text{ when } s = [3, 4] \text{ and } u = 1, \\
	& 1, \text{ when otherwise.} 
	\end{cases}
\end{equation}
The policy in \eqref{eqn:OptimalPolicy_demo} can be obtained using Value Iteration \cite{sutton2018reinforcement} given the true model of the environment. We only present the optimal policy at the critical states because the observational distribution at the critical state-action pairs is only affected by the demonstrator's policy at these critical states.
The demonstrator collects data using an $\epsilon-$greedy policy as described in Section~\ref{sec:ExperienceTransfer}, where $\epsilon = 0.3$. That is, at each state $s$ and observed context $u$, the demonstrator chooses the optimal action in \eqref{eqn:OptimalPolicy_demo} with probability $0.7$ and each of the other three actions with probability $0.1$. If the demonstrator collects data for long enough episodes, by the Law of Large Numbers, the observational distribution of reward-action pairs at state $s$ approaches

\begin{equation}
	\label{eqn:ObservationDistribution_demo}
	P(r, a | s)  = \sum_u P(r | s, a, u) P(a | s, u) P(u).
\end{equation}
We compute the observational distribution $P(r, a | s)$ at the critical positions using \eqref{eqn:ObservationDistribution_demo} by substituting the definition of the reward function, the policy function~\eqref{eqn:OptimalPolicy_demo}, and the distribution of $u$. Next, the learner computes the causal bounds for the rewards at state-action pairs $([0, 3], 1)$, $([1, 4], 4)$, $([3, 4], 2)$ and $([4, 3], 1)$ by solving problem~\eqref{eqn:Optimization_CausalBound} with prior knowledge 
where, except for the rewards of those critical state-action pairs, all other actions have reward $-1$.

 
 \begin{figure}
 	\centering
 	\includegraphics[width = .65\columnwidth]{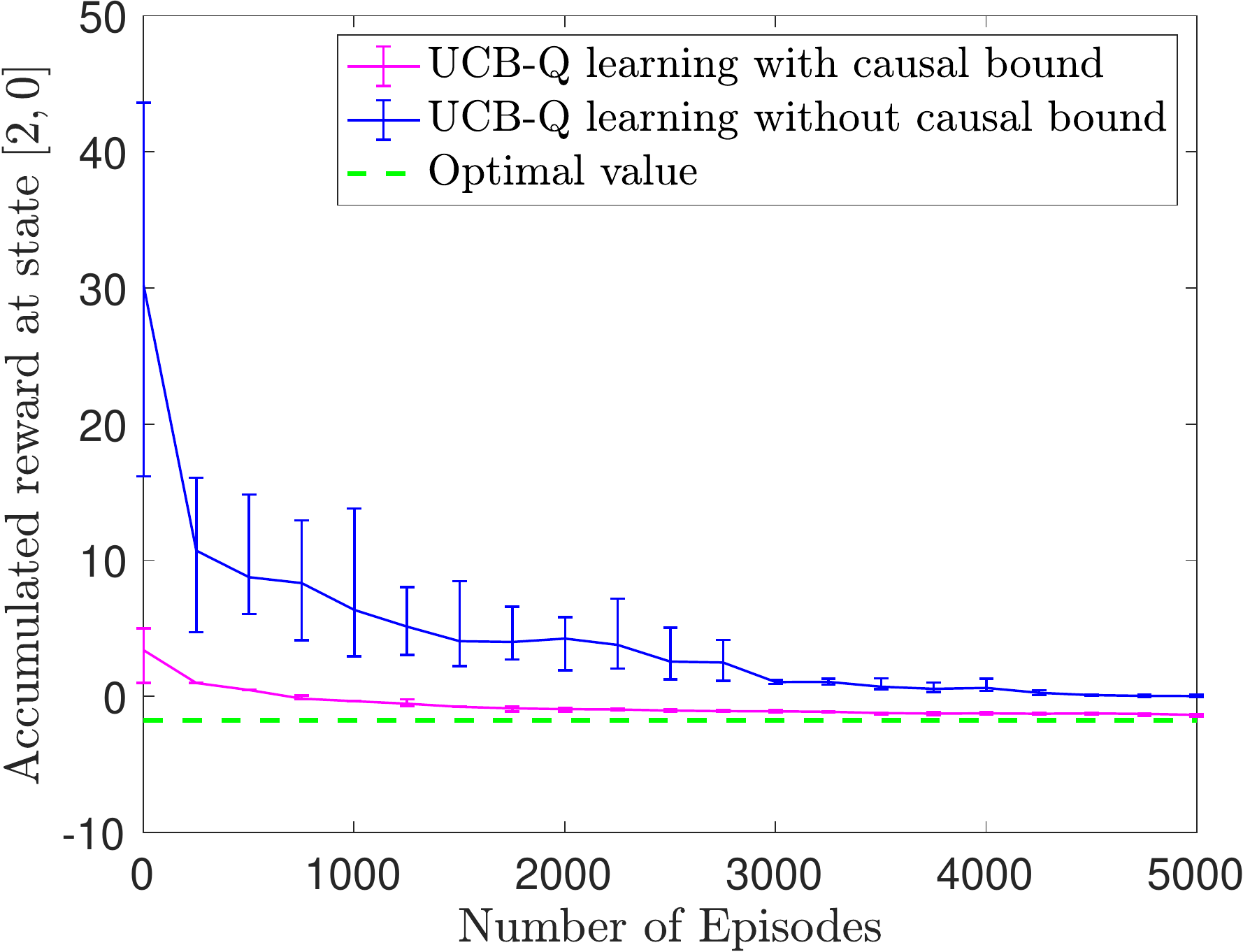}
 	\caption{Learning progress of Algorithm \ref{alg:CB-UCBQ} (magenta curve) compared to UCB-$Q$ learning algorithm \cite{jin2018q} (blue curve) without using causal bounds. The optimal value at state $[2, 0]$ is computed using the true model (green dashed line). Both algorithms are run for $10$ trials and the mean-error curves are presented.}
 	\label{fig:UCBQ_Reward}	
 	\vspace{-2mm}
 \end{figure}
 
 The causal bounds on the rewards at these state-action pairs are listed in Table~\ref{table:CausalBounds_Reward}, together with the true causal effect $\mathbb{E}\big[r | s, do(a)\big]$ and the naive estimation $\mathbb{E}\big[r |s, a \big]$ ignoring the hidden context. At  $([0, 3], 1)$ and $([4, 3], 1)$, the naive estimation returns the correct results. As we show in Section~\ref{sec:ModelBias}, this is because the demonstrator chooses action $a = 1$ with same probability regardless of the context variable $u$. In this case, the naive estimation \eqref{eqn:Prob_X} is equivalent to the causal relationship \eqref{eqn:Prob_doX}. On the other hand, the demonstrator has higher probability to choose action $4$ when $u = 0$ than $u = 1$ at position $[1, 4]$. Then, from Table~\ref{table:CausalBounds_Reward}, we observe that the naive method over-estimates the expected reward at $([1, 4], 4)$. Therefore, if we directly take the reward estimates $\mathbb{E}\big[r |s, a \big]$ in Table~\ref{table:CausalBounds_Reward} and apply Value Iteration \cite{sutton2018reinforcement} to find the optimal value function and policy, we will obtain a suboptimal policy. This is shown in Figure~\ref{fig:Qlearning_wo_causal}. We observe that the naive method gives an over-optimistic value estimate at state $[2, 0]$ (black dashed line) compared to the true value (green dashed line). Nevertheless, the actual performance of the suboptimal policy (red curve) is far below the true optimal performance (green dashed line). This is because the agent over-estimates the reward at $([1, 4], 4)$ and plans to move to $[1, 4]$, while the actual reward it receives is much lower than the reward it would receive if it moved to $[3, 4]$ or $[4, 3]$.
 
 
 Next, we validate Algorithms~\ref{alg:CBC-Q} and \ref{alg:CB-UCBQ} using the bounds on the value functions $[\uline{Q}(s_t, a_t), \overline{Q}(s_t, a_t) ]$ that are computed based on the causal bounds in Table~\ref{table:CausalBounds_Reward}. 
When the causal bounds are only derived at the critical state-action pairs, the bounds on the value functions $[\uline{Q}(s_t, a_t), \overline{Q}(s_t, a_t) ]$ can be found in closed-form, instead of solving the general optimization problems in \eqref{eqn:LP_UpperBound} and \eqref{eqn:LP_LowerBound}.
 The performance of Algorithm~\ref{alg:CBC-Q} is compared to Q learning without demonstrator's data [cf. \eqref{eqn:Qlearning}] and the results are shown in Figure~\ref{fig:Qlearning_wo_causal}. While both Algorithm~\ref{alg:CBC-Q} and plain Q learning converge to the true value, Algorithm~\ref{alg:CBC-Q} requires much fewer data. This is because in plain $Q$ learning (blue curve), the value at state $[2, 0]$
 is estimated incorrectly during the early stages due to the incorrect value estimates at its successor states. By projecting onto the causal bounds using \eqref{eqn:Qlearning_Constrained}, these initial errors in value estimation are controlled and the learning process is accelerated. On the other hand, the performance of Algorithm~\ref{alg:CB-UCBQ} is presented in Figure~\ref{fig:UCBQ_Reward}. Specifically, we compare Algorithm~\ref{alg:CB-UCBQ} with the plain  UCB-$Q$ learning algorithm in \cite{jin2018q}. UCB-$Q$ learning learns the upper confidential bound on the state values and chooses to explore the actions that maximize these bounds. 
 Since the width of the confidential interval in \eqref{eqn:UCBQ} is of order $O(\frac{1}{\sqrt{k}})$ where $k$ is the total number of times a state-action pair is visited, the upper confidential bounds for most state-action pairs are loose at the beginning while these state-action pairs have not been visited. This explains why the value estimate of the UCB-$Q$ learning algorithm (blue curve) in Figure~\ref{fig:UCBQ_Reward} is initially far  from the true value. 
 Furthermore, since the upper confidential bound is loose, UCB-$Q$ explores many low-gain states that have not been visited before and the learning progress of UCB-Q is slow.
 The causal bounds tighten the upper confidential bound, therefore, avoiding unnecessary exploration. As a result, in Figure~\ref{fig:UCBQ_Reward}, we observe that the Causal Bound aided UCB-$Q$ learning algorithm (magenta curve) converges faster than the standard UCB-$Q$ learning algorithm.

\subsection{Unobserved Contextual Transition Function}
\label{sec:Sim_ContextualTransition}

\begin{figure}[t]
	\centering
	\includegraphics[width = .45\columnwidth]{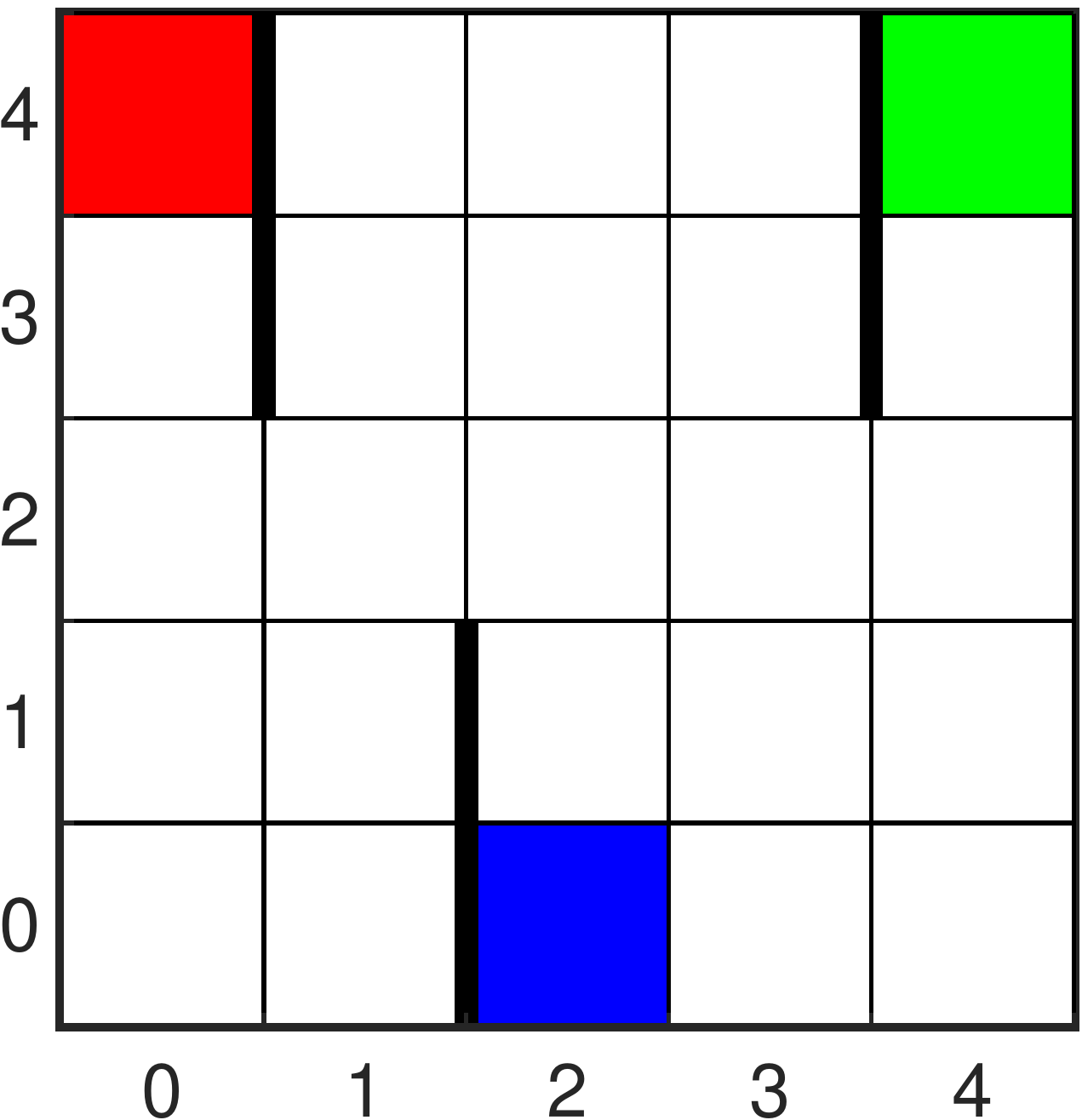}
	\caption{A motion planning problem where the agent starts from the blue cell and moves to the red or green cells. The transition probability at state-action pairs $([0,2], 1)$ and $([4,2], 1)$ are affected by the context variable $u$.}
	\label{fig:field_transition}
\end{figure}

\begin{table}[t]
	\centering
	\begin{tabular}{| c | c | c | c |} 
		\hline
		$(s, a, s')$ & $P\big(s' | s, do(a)\big)$ & $P\big(s' |s, a \big)$ & Causal bounds\\
		\hline
		$([0, 2], 1, [0, 3])$ & $0.18$ & $0.46$ & $\big[0.1020, 0.8820\big]$ \\
		$([0, 2], 1, [0, 1])$ & $0.82$ & $0.54$ & $\big[0.1180, 0.8980\big]$ \\
		$([4, 2], 1, [4, 3])$ & $0.72$ & $0.55$ & $\big[0.12, 0.90\big]$ \\
		$([4, 2], 1, [4, 1])$ & $0.28$ & $0.45$ & $\big[0.10, 0.88\big]$ \\		
		\hline
	\end{tabular}
	\caption{\scriptsize The true transition probability $P\big(s' | s, do(a)\big)$, direct estimate of the probability from the observational distribution $P(r, a|s)$ without considering the hidden contexts and the causal bounds obtained by solving problem~\eqref{eqn:Optimization_CausalBound}. }
	\label{table:CausalBounds_Transition}
	\vspace{-6mm}
\end{table}

\begin{figure}[t]
	\centering
	\includegraphics[width = .7\columnwidth]{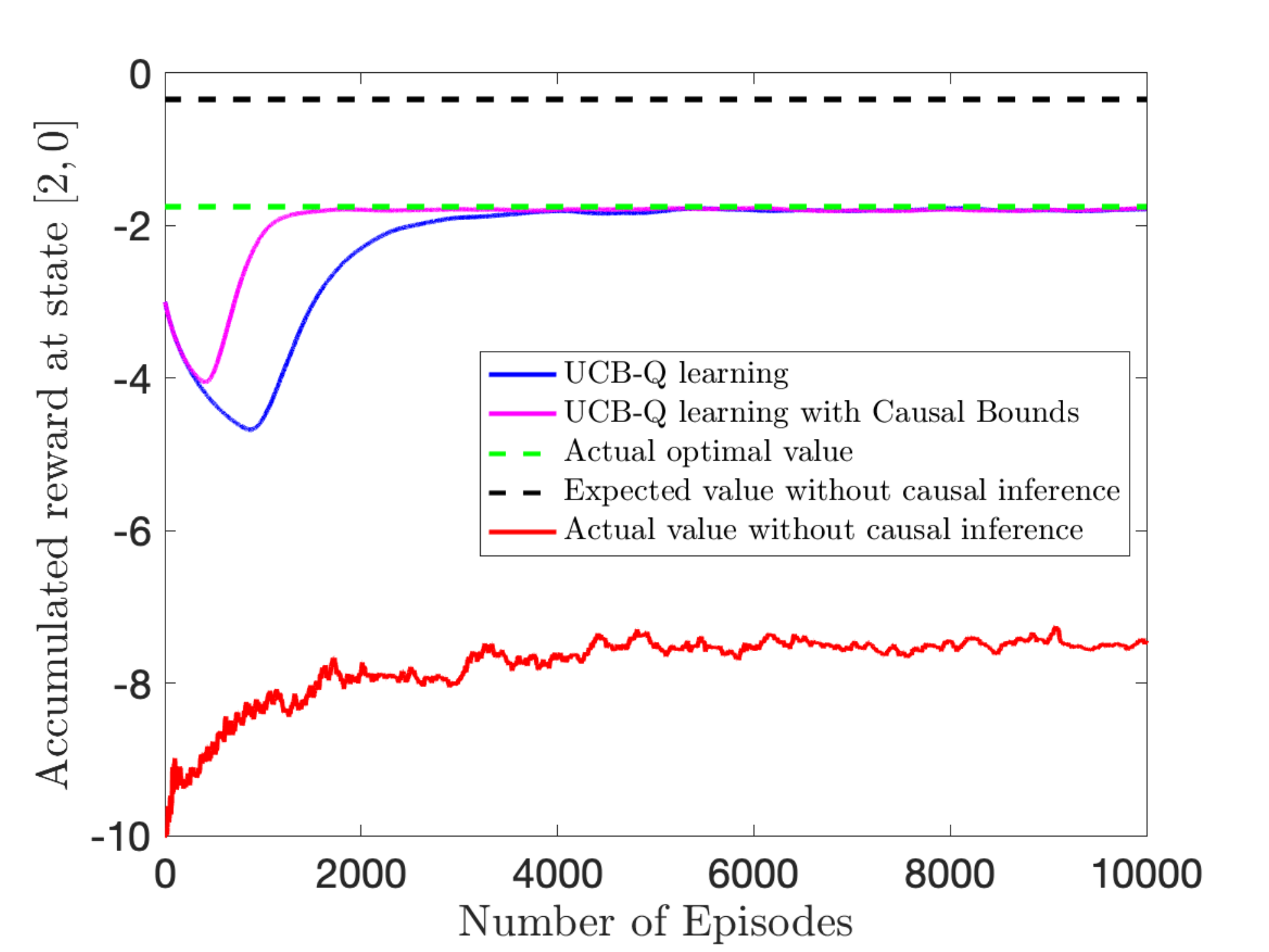}
	\caption{Comparison between the proposed Causal Bound constrained $Q$ learning and the standard $Q$ learning. The blue curve shows the learning progress of the $Q$ learning algorithm \cite{watkins1992q} without using data from the demonstrator. The magenta curve shows the performance of Algorithm~\ref{alg:CBC-Q}. Both curves are obtained by running these algorithms for $10$ trials and take the average. The black (or green) dashed line is the optimal value at state $[2, 0]$ computed using $P\big(s' |s, a \big)$ (or $P\big(s' |s, do(a) \big)$ ). The red curve shows the evaluation of the sub-optimal policy computed using $P\big(s' |s, a \big)$.}
	\label{fig:Qlearning_wo_Causal_Transit}
	\vspace{-2mm}
\end{figure}

Consider the motion planning problem in Figure~\ref{fig:field_transition}, which is the same as the problem considerd in FIgure~\ref{fig:field_Rwd}, except that the rewards received when reaching the red and green cells are now $+10$ and $+5$, respectively, and are not affected by the context variable $u$. Moreover, under context $u = 0$, when the robot takes action $a = 1$ at position $[0,2]$ (or $[4,2]$), it moves upwards with probability $0.7$ (or $0.4$) and moves downwards otherwise. Under context $u = 1$, when the robot takes action $a = 1$ at position $[0,2]$ (or $[4,2]$), it moves upwards with probability $0.05$ (or $0.8$) and moves downwards otherwise.  

Consider a demonstrator agent who knows the optimal contextual policy. Specifically, at the positions $[0,2]$ and $[4,2]$, its optimal policy is

\begin{equation}
\label{eqn:OptimalPolicy_demo_transit}
\pi^\ast(s, u) =
\begin{cases}
& 1, \text{ when } s = [0, 2] \text{ and } u = 0,\\
& 2, \text{ when } s = [0, 2] \text{ and } u = 1,\\
& 4, \text{ when } s = [4, 2] \text{ and } u = 0,\\
& 1, \text{ when } s = [4, 2] \text{ and } u = 1.
\end{cases}
\end{equation}
The above policy is computed using Value Iteration given the true model of the environment. Same as with the policy in \eqref{eqn:OptimalPolicy_demo}, here too we only present the optimal policy at critical states in \eqref{eqn:OptimalPolicy_demo_transit}.
The demonstrator collects samples using an $\epsilon-$greedy policy at positions $[0,2]$ and $[4,2]$, where $\epsilon = 0.3$. The observational distribution of $P(s', a)$ at positions $[0,2]$ and $[4,2]$ can be computed as
\begin{figure}[t]
	\centering
	\includegraphics[width = .7\columnwidth]{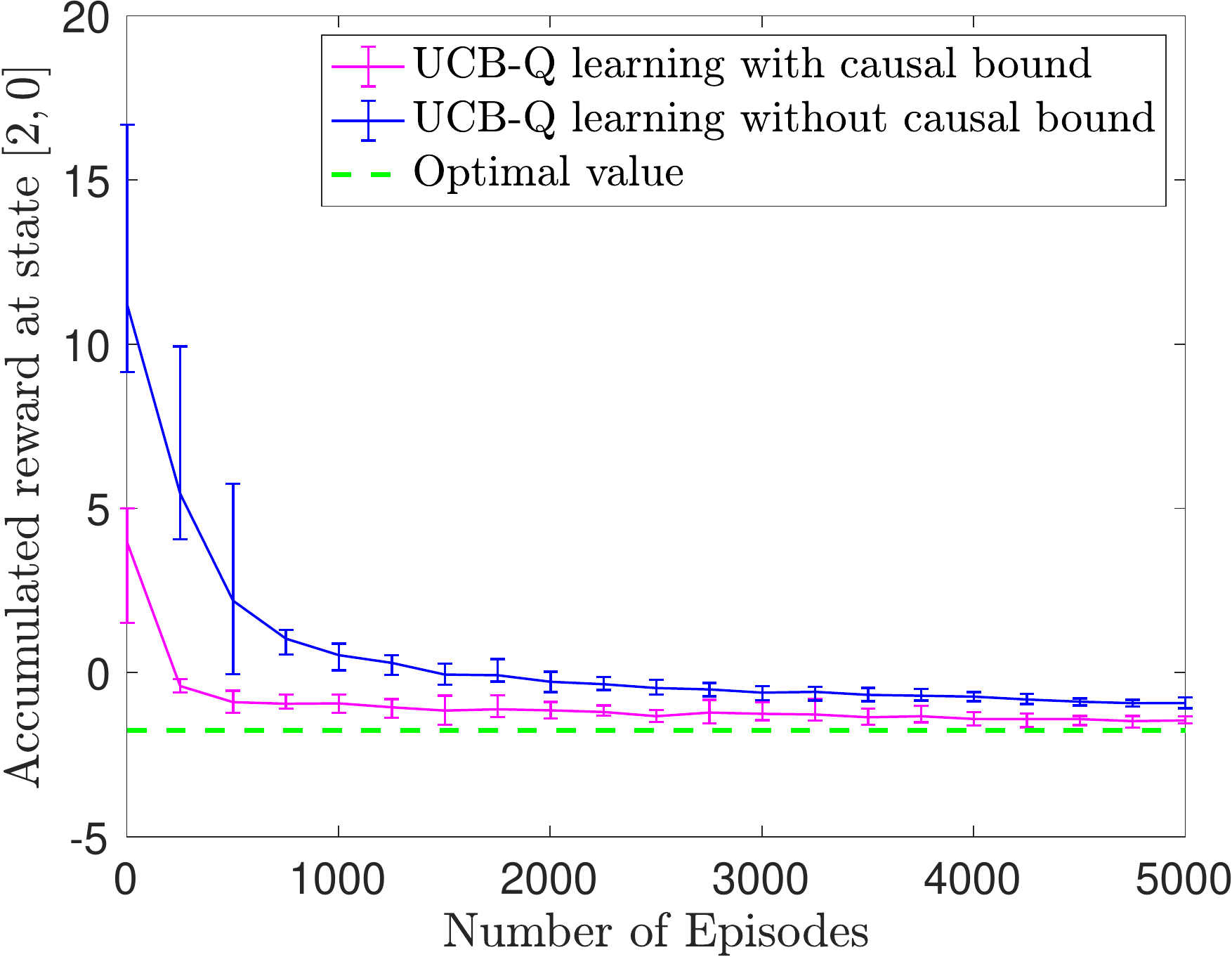}
	\caption{Learning progress of Algorithm \ref{alg:CB-UCBQ} (magenta curve) compared to UCB-$Q$ learning algorithm \cite{jin2018q} (blue curve) without using causal bounds. The optimal value at state $[2, 0]$ is computed using the true model (green dashed line). Both algorithms are run for $10$ trials and the mean-error curves are presented.}
	\label{fig:UCBQ_Transit}
	\vspace{-2mm}
\end{figure}

\begin{equation}
\label{eqn:ObservationDistribution_Transit_demo}
P(s', a | s)  = \sum_u P(s' | s, a, u) P(a | s, u) P(u).
\end{equation}
Given the observational distribution in \eqref{eqn:ObservationDistribution_Transit_demo}, 
we can compute the causal bounds on the transition probabilities at $([0,2], 1)$ and $([4,2], 1)$ using \eqref{eqn:Optimization_CausalBoundTransition}. The causal bounds together with the true transition probabilities and the naive estimation are presented in Table~\ref{table:CausalBounds_Transition}. Since the demonstrator's decisions are affected by the context variable at positions $[0,2]$ and $[4,2]$, we observe that the naive estimation is biased from the true transition probability. 
Figure~\ref{fig:Qlearning_wo_Causal_Transit} shows the suboptimal policy obtained using the naive estimation $P\big(s' |s, a \big)$. We observe that the expected return of the suboptimal policy (black dashed line) is over-optimistic compared to the true optimal return (green dashed line), while the actual return (red curve) we obtain by implementing this suboptimal policy is far lower than the true optimal return. This is because the naive estimation over-estimates the probability that the robot can move upwards to the red cell from position $[0,2]$ by taking action $a = 1$.


Next, we compute the causal bounds on value functions using the causal bounds in Table~\ref{table:CausalBounds_Transition} and use these value function bounds in Algorithms~\ref{alg:CBC-Q} and \ref{alg:CB-UCBQ}. The performance of Algorithm~\ref{alg:CBC-Q} compared to standard $Q$ learning is shown in Figure~\ref{fig:Qlearning_wo_Causal_Transit} and the performance of Algorithm~\ref{alg:CB-UCBQ} compared to the UCB-$Q$ learning method is shown in Figure~\ref{fig:UCBQ_Transit}. The improvement in performance obtained using Algorithms~\ref{alg:CBC-Q} and \ref{alg:CB-UCBQ} is similar to that for the contextual reward case.


\section{Conclusion}
\label{sec:Conclusion}

In this paper, we studied a transfer reinforcement learning problem, where a context-aware demonstrator agent transfers experience in the form of transition and reward samples to a context-unaware learner agent, excluding the contextual information. The goal of the learner is to find a context-unaware optimal policy. We showed how to obtain provable causal bounds on the reward, transition and value functions, and proposed a $Q$ learning and a UCB-$Q$ learning method that employ these causal bounds to reduce the number of samples required by the learner to find the optimal value function without any bias. We provided numerical examples showing the efficacy of the proposed methods.

\newcommand{\BIBdecl}{\setlength{\itemsep}{0.25 em}}
\bibliographystyle{IEEEtran}
\bibliography{biblio}

\begin{thebibliography}{10}
\providecommand{\url}[1]{#1}
\csname url@samestyle\endcsname
\providecommand{\newblock}{\relax}
\providecommand{\bibinfo}[2]{#2}
\providecommand{\BIBentrySTDinterwordspacing}{\spaceskip=0pt\relax}
\providecommand{\BIBentryALTinterwordstretchfactor}{4}
\providecommand{\BIBentryALTinterwordspacing}{\spaceskip=\fontdimen2\font plus
\BIBentryALTinterwordstretchfactor\fontdimen3\font minus
  \fontdimen4\font\relax}
\providecommand{\BIBforeignlanguage}[2]{{%
\expandafter\ifx\csname l@#1\endcsname\relax
\typeout{** WARNING: IEEEtran.bst: No hyphenation pattern has been}%
\typeout{** loaded for the language `#1'. Using the pattern for}%
\typeout{** the default language instead.}%
\else
\language=\csname l@#1\endcsname
\fi
#2}}
\providecommand{\BIBdecl}{\relax}
\BIBdecl

\bibitem{sutton2018reinforcement}
R.~S. Sutton and A.~G. Barto, \emph{Reinforcement learning: An
  introduction}.\hskip 1em plus 0.5em minus 0.4em\relax MIT press, 2018.

\bibitem{taylor2009transfer}
M.~E. Taylor and P.~Stone, ``Transfer learning for reinforcement learning
  domains: A survey,'' \emph{Journal of Machine Learning Research}, vol.~10,
  no. Jul, pp. 1633--1685, 2009.

\bibitem{taylor2007transfer}
M.~E. Taylor, P.~Stone, and Y.~Liu, ``Transfer learning via inter-task mappings
  for temporal difference learning,'' \emph{Journal of Machine Learning
  Research}, vol.~8, no. Sep, pp. 2125--2167, 2007.

\bibitem{barreto2017successor}
A.~Barreto, W.~Dabney, R.~Munos, J.~J. Hunt, T.~Schaul, H.~P. van Hasselt, and
  D.~Silver, ``Successor features for transfer in reinforcement learning,'' in
  \emph{Advances in neural information processing systems}, 2017, pp.
  4055--4065.

\bibitem{finn2017model}
C.~Finn, P.~Abbeel, and S.~Levine, ``Model-agnostic meta-learning for fast
  adaptation of deep networks,'' in \emph{Proceedings of the 34th International
  Conference on Machine Learning-Volume 70}.\hskip 1em plus 0.5em minus
  0.4em\relax JMLR. org, 2017, pp. 1126--1135.

\bibitem{konidaris2012transfer}
G.~Konidaris, I.~Scheidwasser, and A.~Barto, ``Transfer in reinforcement
  learning via shared features,'' \emph{Journal of Machine Learning Research},
  vol.~13, no. May, pp. 1333--1371, 2012.

\bibitem{lazaric2008transfer}
A.~Lazaric, M.~Restelli, and A.~Bonarini, ``Transfer of samples in batch
  reinforcement learning,'' in \emph{Proceedings of the 25th international
  conference on Machine learning}.\hskip 1em plus 0.5em minus 0.4em\relax ACM,
  2008, pp. 544--551.

\bibitem{lazaric2011transfer}
A.~Lazaric and M.~Restelli, ``Transfer from multiple mdps,'' in \emph{Advances
  in Neural Information Processing Systems}, 2011, pp. 1746--1754.

\bibitem{taylor2008transferringModel}
M.~E. Taylor, N.~K. Jong, and P.~Stone, ``Transferring instances for
  model-based reinforcement learning,'' in \emph{Joint European Conference on
  Machine Learning and Knowledge Discovery in Databases}.\hskip 1em plus 0.5em
  minus 0.4em\relax Springer, 2008, pp. 488--505.

\bibitem{tirinzoni2018importance}
A.~Tirinzoni, A.~Sessa, M.~Pirotta, and M.~Restelli, ``Importance weighted
  transfer of samples in reinforcement learning,'' in \emph{International
  Conference on Machine Learning}, 2018, pp. 4943--4952.

\bibitem{tirinzoni2019transfer}
A.~Tirinzoni, M.~Salvini, and M.~Restelli, ``Transfer of samples in policy
  search via multiple importance sampling,'' in \emph{International Conference
  on Machine Learning}, 2019, pp. 6264--6274.

\bibitem{rosenstein2005transfer}
M.~T. Rosenstein, Z.~Marx, L.~P. Kaelbling, and T.~G. Dietterich, ``To transfer
  or not to transfer,'' in \emph{NIPS 2005 workshop on transfer learning}, vol.
  898, 2005, p.~3.

\bibitem{hallak2015contextual}
A.~Hallak, D.~Di~Castro, and S.~Mannor, ``Contextual markov decision
  processes,'' \emph{arXiv preprint arXiv:1502.02259}, 2015.

\bibitem{pearl2009causality}
J.~Pearl, \emph{Causality: models, reasoning and inference}, 2nd~ed.\hskip 1em
  plus 0.5em minus 0.4em\relax Cambridge University Press, New York, 2009.

\bibitem{watkins1992q}
C.~J. Watkins and P.~Dayan, ``Q-learning,'' \emph{Machine learning}, vol.~8,
  no. 3-4, pp. 279--292, 1992.

\bibitem{azar2017minimax}
M.~G. Azar, I.~Osband, and R.~Munos, ``Minimax regret bounds for reinforcement
  learning,'' in \emph{Proceedings of the 34th International Conference on
  Machine Learning}, 2017, pp. 263--272.

\bibitem{jin2018q}
C.~Jin, Z.~Allen-Zhu, S.~Bubeck, and M.~I. Jordan, ``Is q-learning provably
  efficient?'' in \emph{Advances in Neural Information Processing Systems},
  2018, pp. 4863--4873.

\bibitem{dong2019q}
K.~Dong, Y.~Wang, X.~Chen, and L.~Wang, ``Q-learning with ucb exploration is
  sample efficient for infinite-horizon mdp,'' \emph{arXiv preprint
  arXiv:1901.09311}, 2019.

\bibitem{zhang2017transfer}
J.~Zhang and E.~Bareinboim, ``Transfer learning in multi-armed bandits: a
  causal approach,'' in \emph{Proceedings of the 26th International Joint
  Conference on Artificial Intelligence}.\hskip 1em plus 0.5em minus
  0.4em\relax AAAI Press, 2017, pp. 1340--1346.

\bibitem{auer2002finite}
P.~Auer, N.~Cesa-Bianchi, and P.~Fischer, ``Finite-time analysis of the
  multiarmed bandit problem,'' \emph{Machine learning}, vol.~47, no. 2-3, pp.
  235--256, 2002.

\bibitem{jaakkola1995reinforcement}
T.~Jaakkola, S.~P. Singh, and M.~I. Jordan, ``Reinforcement learning algorithm
  for partially observable markov decision problems,'' in \emph{Advances in
  neural information processing systems}, 1995, pp. 345--352.

\bibitem{hausknecht2015deep}
M.~Hausknecht and P.~Stone, ``Deep recurrent q-learning for partially
  observable mdps,'' in \emph{2015 AAAI Fall Symposium Series}, 2015.

\bibitem{tu2018least}
S.~Tu and B.~Recht, ``Least-squares temporal difference learning for the linear
  quadratic regulator,'' in \emph{International Conference on Machine
  Learning}, 2018, pp. 5012--5021.

\bibitem{sun2019model}
W.~Sun, N.~Jiang, A.~Krishnamurthy, A.~Agarwal, and J.~Langford, ``Model-based
  rl in contextual decision processes: Pac bounds and exponential improvements
  over model-free approaches,'' in \emph{Conference on Learning Theory}, 2019,
  pp. 2898--2933.

\bibitem{d1963probabilistic}
F.~d'Epenoux, ``A probabilistic production and inventory problem,''
  \emph{Management Science}, vol.~10, no.~1, pp. 98--108, 1963.

\bibitem{bertsimas1997introduction}
D.~Bertsimas and J.~N. Tsitsiklis, \emph{Introduction to linear
  optimization}.\hskip 1em plus 0.5em minus 0.4em\relax Athena Scientific
  Belmont, MA, 1997, vol.~6.

\bibitem{kearns2002near}
M.~Kearns and S.~Singh, ``Near-optimal reinforcement learning in polynomial
  time,'' \emph{Machine learning}, vol.~49, no. 2-3, pp. 209--232, 2002.

\end{thebibliography}

\newpage
\section*{Appendix A}
\subsection*{Proof of Theorem~\ref{thm:Bd_ValueFunc}}
\label{sec:Proof}

\begin{algorithm}[t]
	\small
	\caption{Computing bounds on $Q(s, a)$}\label{alg:Q_bound}
	\KwIn{The bounds on the reward, transition and state-based value function $[\uline{r}(s, a), \overline{r}(s, a)]$, $[\;\uline{P}(s'|s, a), \overline{P}(s'|s, a) \;]$ and $[\; \uline{V}(s), \bar{V}(s) \;]$. }
	\For{ every $(s, a) \in \mathcal{S} \times \mathcal{A}$}{
		Set $\mathcal{S}' \leftarrow$ $\{$all possible next states $s'$ at state $s$ when action $a$ is taken$\}$  \;
		$p, q \leftarrow 0 \in \mathbb{R}^{|\mathcal{S}'|}$ \;
		$\vec{v} \leftarrow [V(s_1'), V(s_2'), \dots, V(s_{|S'|}')]^T$ so that $V(s_1') \geq V(s_2') \geq \dots \geq V(s_{|S'|}')$ \;
		$p(1)  \leftarrow \overline{P}(s_1' | s, a)$, $q(|S'|) \leftarrow \overline{P}(s_{|S'|}' | s, a)$ \;
		$M \leftarrow \overline{P}(s_1' | s, a)$, $N \leftarrow \overline{P}(s_{|S'|}' | s, a)$\;
		\For{$i = 2, 3, \dots, |S'|$}{
			\If{$1 - M \leq \overline{P}(s_i' | s, a)$,}{
				$p(i) \leftarrow 1 - M$\;
				$p(j) \leftarrow 0$ for all $i < j \leq |S'|$ \;
				break \;
			}
			\Else{$p(i) \leftarrow \overline{P}(s_i' | s, a)$ \;
				$M \leftarrow M + P(s_i' | s, a)$ \;
			}
			\If{$1 - N \leq \overline{P}(s_{|S'|+1-i}' | s, a)$,}{
				$p(|S'|+1-i) \leftarrow 1 - N$\;
				$p(j) \leftarrow 0$ for all $1 \leq j < |S'|+1-i$ \;
				break \;
			}
			\Else{$p(i) \leftarrow \overline{P}(s_{|S'|+1-i}' | s, a)$ \;
				$N \leftarrow N + P(s_{|S'|+1-i}' | s, a)$ \;
			}
		}
		$\overline{Q}(s, a) \leftarrow \overline{r}(s, a) + \gamma p^T \vec{v}$ \;
		$\uline{Q}(s, a) \leftarrow \uline{r}(s, a) + \gamma q^T \vec{v}$.
	}
\end{algorithm}

	First, we show the bound in \eqref{eqn:UpBd_Value}. Since $\{ V^\ast(s) \}$ is the solution to problem~\eqref{eqn:LP_UpperBound}, we obtain that
	
	\begin{equation}
	\label{eqn:Inequality_1}
	\sum_{s \in \mathcal{S} } c(s) V^\ast(s) \geq \sum_{s \in \mathcal{S} } c(s) \bar{V}^\ast(s).
	\end{equation}
	Subtracting $\sum_{ \tilde{s} \neq s } c(\tilde{s}) \bar{V}^\ast (\tilde{s}) $ from both sides of \eqref{eqn:Inequality_1}, and dividing by $c(s)$, we get that
	
	\begin{equation}
	\label{eqn:UpBd_Value_1}
	\bar{V}^\ast(s) \leq V^\ast(s) + \frac{1}{c(s)} \big(\sum_{ \tilde{s} \neq s } c(\tilde{s}) V^\ast(\tilde{s})  - \sum_{ \tilde{s} \neq s } c(\tilde{s}) \bar{V}^\ast (\tilde{s})   \big).
	\end{equation}
	Since $r(s, a) \geq \uline{R}$ for all $(s, a)$, we have that any value function satisfies $V(s) = \mathbb{E} \big[ \sum_{t=0}^{\infty} \gamma^t r(s_t, a_t) | s_0 = s \big] \geq \sum_{t=0}^{\infty} \gamma^t \uline{R} = \frac{\uline{R}}{1 - \gamma}$. Replacing $\bar{V}^\ast(\tilde{s})$ in \eqref{eqn:UpBd_Value_1} with $\frac{\uline{R}}{1 - \gamma}$, we achieve the bound in \eqref{eqn:UpBd_Value}.	
	Similary, to show the bound in \eqref{eqn:LwrBd_Value}, since $\{ V^\ast(s) \}$ is the solution to problem~\eqref{eqn:LP_LowerBound}, we obtain that 
	
	\begin{equation}
	\label{eqn:Inequality_2}
	\sum_{s \in \mathcal{S} } c(s) V^\ast(s) \leq \sum_{s \in \mathcal{S} } c(s) \uline{V}^\ast(s)  
	\end{equation}	
	Subtracting $\sum_{ \tilde{s} \neq s } c(\tilde{s}) \uline{V} (\tilde{s}) $ from both sides of \eqref{eqn:Inequality_2}, and dividing by $c(s)$, we get
	
	\begin{equation}
	\label{eqn:LwrBd_Value_1}
	\uline{V}^\ast(s) \geq V^\ast(s) - \frac{1}{c(s)} \big(\sum_{ \tilde{s} \neq s } c(\tilde{s}) \uline{V}^\ast (\tilde{s})  - \sum_{ \tilde{s} \neq s } c(\tilde{s}) V^\ast(\tilde{s}) \big).
	\end{equation}
	Since $r(s, a) \leq \bar{R}$ for all $(s, a)$, we have that any value function satisfies $V(s) = \mathbb{E} \big[ \sum_{t=0}^{\infty} \gamma^t r(s_t, a_t) | s_0 = s \big] \leq \sum_{t=0}^{\infty} \gamma^t \bar{R} = \frac{\bar{R}}{1 - \gamma}$. Replacing $\uline{V}(\tilde{s})$ in \eqref{eqn:LwrBd_Value_1} with $\frac{\bar{R}}{1 - \gamma}$, we achieve the bound in \eqref{eqn:LwrBd_Value}.

\section*{Appendix B}
\subsection*{Computation of Bounds on $Q(s, a)$}
\label{sec:ComputationQ}
In this section, we present an algorithm to compute the upper and lower bounds on the action-based value function $Q(s, a)$ using the causal bounds $[\uline{r}(s, a), \overline{r}(s, a)]$, $[\uline{P}(s'|s, a), \overline{P}(s'|s, a) ]$ and $[\uline{V}(s), \bar{V}(s)]$ derived in Section~\ref{sec:CausalBound_ValueFunctions}. This algorithm is illustrated in Algorithm~\ref{alg:Q_bound}.

\end{document}